%% file: main.tex
\documentclass{article}

\usepackage{microtype}
\usepackage{graphicx}
\usepackage{subfigure}
\usepackage{booktabs} 

\usepackage[accepted]{icml2024}

\usepackage{amsmath}

\usepackage{amssymb}
\usepackage{mathtools}
\usepackage{amsthm}
\usepackage{hyperref}
\usepackage{url}
\usepackage{tabularx}
\usepackage{booktabs}
\usepackage{multirow}
\usepackage{hhline}
\usepackage{array, makecell}
\usepackage{graphicx}
\usepackage{mathrsfs}  
\usepackage{enumitem}
\usepackage{tikz}
\usepackage{diagbox}
\usepackage{makecell}
\usepackage[ruled,vlined,algo2e]{algorithm2e}
\newcolumntype{C}[1]{>{\centering\arraybackslash}p{#1}}
\usepackage[capitalize,noabbrev]{cleveref}
\theoremstyle{plain}

\theoremstyle{definition}

\theoremstyle{remark}

\newcommand\myeq{\mkern3.0mu{=}\mkern3.0mu}
\usepackage[textsize=tiny]{todonotes}
\usepackage{wrapfig}

\icmltitlerunning{Representing Molecules as Random Walks Over Interpretable Grammars}

\begin{document}

\twocolumn[
\icmltitle{Representing Molecules as Random Walks Over Interpretable Grammars}




\begin{icmlauthorlist}
\icmlauthor{Michael Sun}{csail}
\icmlauthor{Minghao Guo}{csail}
\icmlauthor{Weize Yuan}{mitchem}
\icmlauthor{Veronika Thost}{IBM}
\icmlauthor{Crystal Elaine Owens}{csail}
\icmlauthor{Aristotle Franklin Grosz}{mitcheme}
\icmlauthor{Sharvaa Selvan}{mit}
\icmlauthor{Katelyn Zhou}{well}
\icmlauthor{Hassan Mohiuddin}{mit}
\icmlauthor{Benjamin J Pedretti}{mitcheme}
\icmlauthor{Zachary P Smith}{mitcheme}
\icmlauthor{Jie Chen}{IBM}
\icmlauthor{Wojciech Matusik}{csail}
\end{icmlauthorlist}

\icmlaffiliation{csail}{MIT CSAIL}
\icmlaffiliation{mit}{MIT}
\icmlaffiliation{well}{Wellesley}
\icmlaffiliation{mitchem}{MIT Chemistry}
\icmlaffiliation{mitcheme}{MIT Chemical Engineering}
\icmlaffiliation{IBM}{MIT-IBM Watson AI Lab, IBM Research}

\icmlcorrespondingauthor{Michael Sun}{msun415@csail.mit.edu}

\icmlkeywords{Machine Learning, ICML}

\vskip 0.3in
]



\printAffiliationsAndNotice{}  

\begin{abstract}
Recent research in molecular discovery has primarily been devoted to small, drug-like molecules, leaving many similarly important applications in material design without adequate technology. These applications often rely on more complex molecular structures with fewer examples that are carefully designed using known substructures. We propose a data-efficient and interpretable model for representing and reasoning over such molecules in terms of graph grammars that explicitly describe the hierarchical design space featuring motifs to be the design basis. We present a novel representation in the form of random walks over the design space, which facilitates both molecule generation and property prediction. We demonstrate clear advantages over existing methods in terms of performance, efficiency, and synthesizability of predicted molecules, and we provide detailed insights into the method's chemical interpretability. Code is available at \url{https://github.com/shiningsunnyday/polymer_walk}.
\end{abstract}

\input{./body/intro-vt}

\input{./body/1.introduction}
\input{./body/2.relatedworks}

\input{./body/3.method}

\input{./body/4.results}

\input{./body/5.discussion_analysis}

\section*{Acknowledgements}

This work is supported by the MIT-IBM Watson AI Lab and its member companies Evonik and Shell.

\section*{Impact Statement}

This paper presents work whose goal is to concurrently and conjointly advance the fields of 
Machine Learning and Chemical Discovery. The application of our method can have consequences for real-world discovery workflows. There are no ethical aspects which we foresee and feel must be discussed here.


\bibliography{main}
\bibliographystyle{icml2024}

\newpage
\appendix
\onecolumn
\input{./body/6.appendix}

\end{document}

%% file: body/intro-vt.tex
\section{Introduction}

Property-driven molecular discovery represents a challenging application with great potential benefits for society, and this is reflected in the large amount of research conducted in the machine learning community on this topic in recent years \cite{some-survey}.
Yet, most of the research focuses on small, drug-like molecules, while many classes of more complex molecules have been largely neglected.
Materials designed for applications such as gas-separation membranes or photovoltaics, which are critical for a sustainable future, often have specific distributions of molecule structure that differ significantly from typical drug-like molecules. In addition, the specificity of the designs and use cases, and the considerable cost of practical experiments, make it often a scenario that is scarce in both data and labels; for example, datasets of $\approx 300$ molecules or less are not uncommon \cite{microporous,hopv,Helma01}. As a consequence, materials science has not yet fully exploited the potential of machine learning methods
\cite{rf1, rf2}.
We focus on such challenging
datasets that feature complex molecules containing functional groups and structural motifs which are applied in multiple diverse, real-world application scenarios.

Our goal is to represent and reason about molecules in a data-efficient and interpretable way. Domain-specific datasets typically exhibit distinct motifs and functional groups, which serve as structural priors in our molecular representation. Previous works show that structural priors are highly advantageous for applications that require data efficiency \cite{ecfp,2,hmgnn,4,5}. We propose a novel approach to molecular discovery that is tailored to more complex molecules and low-data scenarios and builds upon the above insights. The idea is to start from a set of expert-defined motifs\footnote{Note that our method works with any given set of motifs (e.g., we can apply one of the more simple algorithms used in existing works), but our evaluation shows that certain applications benefit from high-quality domain knowledge.} and learn a context-sensitive grammar over the space of motifs. The \emph{novelty of this work lies in our representation and learning of this grammar}.

We define a \emph{motif graph} -- a hierarchical abstraction of the molecular design space induced by the given data, where each node is a motif and each edge represents a possible attachment between a pair of motifs. 
Our main technical contribution is an efficient and interpretable parameterization over the context-sensitive grammar induced by the design space, and the description of a molecule as a random walk of context-sensitive transition rules. Our representation of molecules combines the quality of representation learning with the interpretability of a rule-based grammar. 

In terms of quality, we demonstrate our grammar representation suits applications characterized by designer molecules. We select datasets that reflect real-world settings of experimentally curated designs of molecules with complex, modular sub-structures characterized by functional groups known or hypothesized to yield high target properties. 

In terms of interpretability, our grammar representation is special in two ways. As an indirect consequence of supervised learning, our model produces visually discernible clusters according to distinctive structural features within the dataset. More importantly, our compact, context-sensitive \emph{grammar allows for discovering design rules} that reveal the design principles used during the creation of the dataset. 



\begin{itemize}[leftmargin=*,topsep=0pt,noitemsep]
\item Our method largely outperforms pretrained and traditional methods for molecular property prediction. It is competitive with a state-of-the-art graph grammar system for chemistry \cite{ours} in terms of quality while being an order of magnitude more runtime efficient.
\item
 Our method's interpretable representations reveal deeper insights into relationships implicit in the data, explain the model's reasoning, and lead to novel scientific insights.
\item
Our method produces promising 
molecule generations, in particular, producing diverse designs that are synthesizable at a significantly higher rate than the state-of-the-art data-efficient generative model, DEG \cite{deg}. 
\item Finally, made possible by our method's interpretability, our approach enables close collaboration with domain experts. In particular, we devised and executed feasible, practical, and semi-automated workflows with experts
for fragmenting molecules, constructing the design space, and interpreting the results. 
\end{itemize}

%% file: body/1.introduction.tex

%% file: body/2.relatedworks.tex
\section{Related Works}
\textbf{Motif-based molecular property prediction.} 
ECFP embeddings \cite{ecfp}, which capture relevant 
ego-graphs present in a molecule in bit vectors, represent a motif-based encoding. 
ECFP embeddings in combinations with simple predictors (e.g., XGBoost) have been competitive on small datasets \cite{2}. 
In our evaluation, we show that our model is similarly \emph{data-efficient} but delivers a better predictive performance, owing to the use of graph-based representations.
In light of the good performance of ECFPs, it is not surprising that the recently developed subgraph graph neural networks (GNNs) 
report competitive performance in molecular property prediction when using ego-graphs as subgraphs \cite{3}; we consider ESAN \cite{esan} in our evaluation. 
However, existing models usually apply subgraphs rooted at all individual nodes rather than a set of more coarse-grained, potentially complex, domain-specific subgraphs. 
Other recent work that integrates motifs to improve out-of-distribution detection similarly lacks this dimension of modeling \cite{5}.

A few closely related works have recently proposed molecular graph representations where the relations between motifs are explicitly represented,  together with corresponding models \cite{hmgnn, 4}.
Our work is different from theirs in two aspects. First, we show that commonly used automatic approaches for motif extraction are not sufficient for property prediction over several kinds of more complex molecules, and that custom motifs given by domain experts yield better performance. It allows for biasing the model towards known structure-activity relationships or the expert's hypotheses (e.g., fragments known or assumed to be critical for the property under consideration).
Second, to the best of our knowledge, their motif graph representations do not model the context sensitivity explicitly (e.g.,  HM-GNN's motif graph \cite{hmgnn} connects two motifs based on co-occurence in a molecule~only). 

\paragraph{Molecule representation by grammars.}
Recent work has shown that such grammars represent a data-efficient way for representing molecules and yield SOTA results \cite{deg,ours}. In a nutshell, this is achieved by explicitly representing the training data's design space in terms of learnt motifs, in the form of a graph grammar.
\emph{Grammars naturally allow for generating novel molecules in the given design space}. Yet, obtaining production rules involves either manual definition \cite{Krenn19, polygrammar, Nigam21} or a significant complexity to automatically learn \cite{deg, Kajino19}, where the training times for downstream tasks are considerable (see Figure~\ref{size-ablation}). Further, the learnt substructures 
sometimes lack a chemical interpretation, and grammar derivations often produce chemically invalid structures \cite{deg}, so the natural potential of symbolic methods for interpretability and validity is lost, although such elements are critical for 
expert validation
and for gaining scientific insights. 
\emph{We propose a novel way for representing and learning such context-sensitive grammars}, over a design space informed by chemical motifs. This approach results in order-of-magnitude differences in runtime and enhances chemical interpretability.

\paragraph{Other works for molecular representation learning.}
There are various other non-motif based approaches that we compare to in our evaluation, including (pre-trained) GNNs \cite{pretrainGIN,xia2023systematic}, motif-based pre-training approaches designed for semi- or unsupervised learning \cite{xia2023systematic}, and molecular few-shot learning including the SOTA, which relies on modeling the domain expert's reasoning in terms of related molecule contexts using associative memories \cite{7}. Central to our method is the connection between random walks and graph diffusion, established methods that have been particularly effective to model  graph structures through physics-inspired processes \cite{thanou17}. Other related works and more detailed discussions 
can be found in Appendix~\ref{app:C}.

%% file: body/3.method.tex
\section{An Interpretable, Grammar-based Molecule Representation and Efficient Learning}

\begin{figure*}[t]
  \centering
  \includegraphics[width=0.95\linewidth]{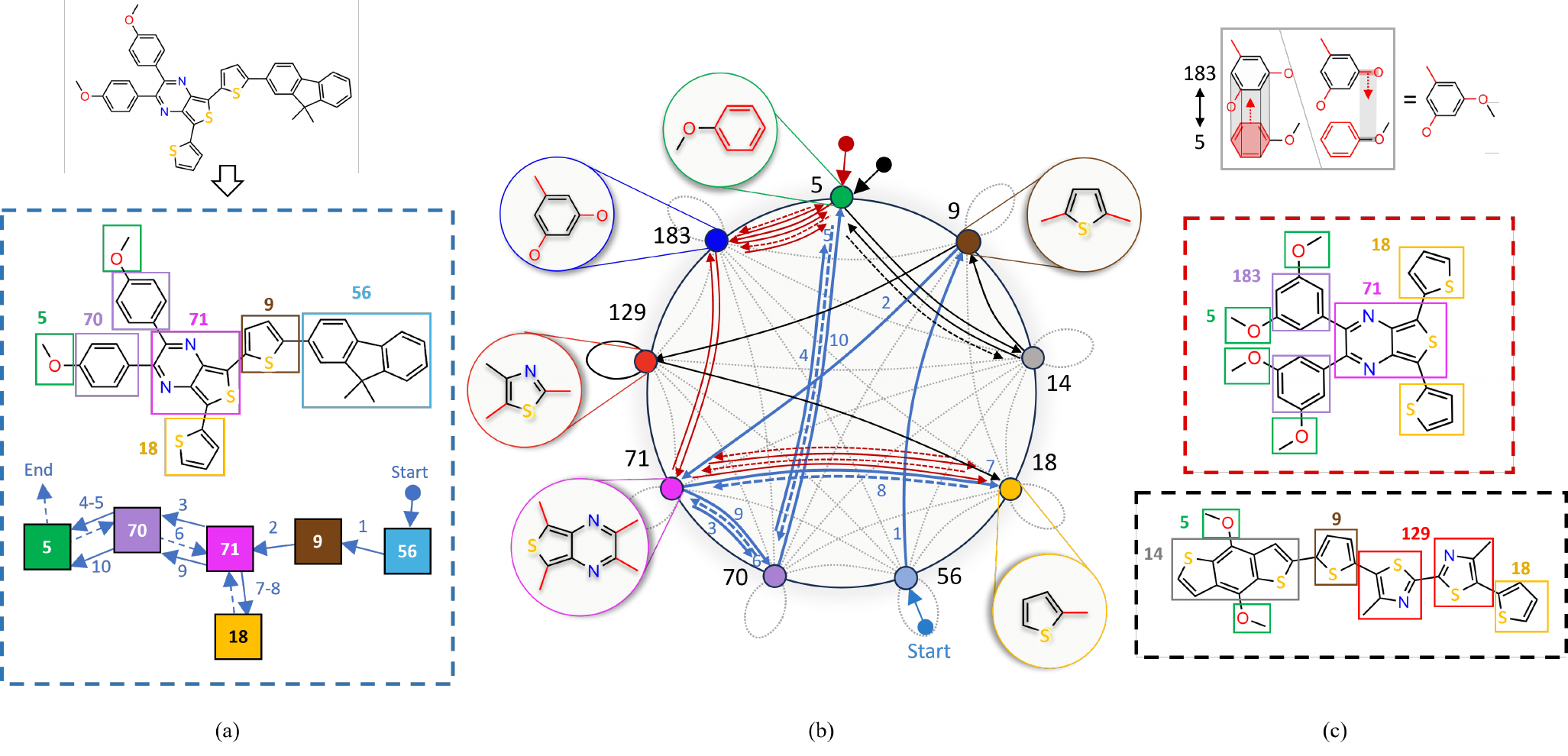}
  \caption{Illustration of our random walk representation: (a) (top) molecule $M$, number 33 (middle) $H_M$ as a connected subgraph of $G$ (bottom) $\hat{H}_M$ as a random walk over $H_M$; (b) the motif graph $G$, each node is a motif $v$ that contains both the molecular fragment $v_B$ (black molecule sections) and the contexts for attachment (\textcolor{red}{$v_R$}, red molecule sections), each \textcolor{gray}{gray} line indicates a possible attachment between nodes. Directed edges of $\hat{H}_M$ use the same color as the dashed border of the corresponding figure of $M$; (c) (top) demonstration of motif matching criteria eq~\ref{eq:criteria-1}-\ref{eq:criteria-4} ($183\leftrightarrow 5$), another example is in Fig.~\ref{match-example} (bottom) two more examples of $H_M$.}
  \label{fig:methodG}
\end{figure*}



Our method employs 
a \emph{graph grammar}, which is composed of a set of predefined molecular motifs and a set of transition rules. The motifs are devised either through automatic generation or manual curation and are interconnected following the transition rules to assemble into a complete molecular structure.
Following~\citep{deg,ours}, a grammar $\mathcal{G}\myeq(\mathcal{N}, \Sigma, \mathcal{P}, \mathcal{X})$ contains a set $\mathcal{N}$ of non-terminal nodes, a set $\Sigma$ of terminal nodes representing chemical atoms, and a starting node $\mathcal{X}$. 
The generation of molecular graphs is described using a set of production rules, $\mathcal{P} = \{p_i | i = 1, \ldots, k\}$. Each rule, $p_i$, is defined by a transformation from a left-hand side ($LHS$) to a right-hand side ($RHS$), with both sides being graphs. The process starts from an initial empty graph $\mathcal{X}$, and a molecule is constructed by iteratively applying a rule from $\mathcal{P}$, where the $LHS$ of the selected rule matches a subgraph within the current graph. This selected subgraph is then replaced by the corresponding $RHS$ of the rule.
\paragraph{Random Walk Grammar.}
We introduce random walk grammar, characterized by a specific condition where the $LHS$ of each rule differs from its $RHS$ by exactly a motif. Such a design ensures that the generation of a molecule is a progressive process, where in each step, a new subgraph is attached to the existing graph. We implement the grammar using a compact \emph{motif graph}~$G$ (Fig.~\ref{fig:methodG}~(b)); the nodes are the motifs and each edge describes the application of a transition rule. 

We highlight two novelties of this work:
\begin{enumerate}[leftmargin=*,topsep=0pt,noitemsep]
    \item Molecules are represented as random walks over connected subgraphs of $G$ (Fig.~\ref{fig:methodG}~(a)). This representation is explicit, compact, and interpretable.
    \item The context-sensitive grammar over $G$ is learnable from a given training dataset by optimizing parameters that determine the prior and adjusted edge weights of $G$. These weights parameterize the transition probabilities, thereby influencing the molecular representation and facilitating the learning of context-sensitive rules, which we elucidate in our analysis section.
\end{enumerate}

We demonstrate the utility of our grammar-based representation for both molecular generation and property prediction tasks. The main steps of our workflow are as follows.


\paragraph{Motif-based Molecule Fragmentation.}
Our method builds upon a given molecule fragmentation.
More specifically, given a dataset 
$D=\{M^{(i)}:=(V_{M^{(i)}}, E_{M^{(i)}})\}_ {i=1}^{|D|}$, a fragmentation of $M^{(i)}$ is a collection of disjoint molecular graphs $\{F^{(i)}_j\} := \{(V^{(i)}_j, E^{(i)}_j)\}$ such that $\sqcup_j V^{(i)} = V_{M^{(i)}}$. Letting $g(v)$ denote the node-induced subgraph of $g$ by $v$, $F^{(i)}_j$ is the subgraph of $M^{(i)}$ induced by $V^{(i)}_j$. When  $F^{(i)}_j$ is a chemical motif, it is essential to know the possible contexts within which $F^{(i)}_j$ occur, because the behavior of one substructure is often influenced by neighboring structures\footnote{For materials applications that rely in particular on electrophilicity, polarity, and extended aromaticity, longer-range combinations and patterns of motifs are often more influential than any individual one.\label{footnote:2}}. Specifically, given neighboring fragments $j_1, j_2$, i.e. $ \exists e \in E_{M^{(i)}} \text{ s.t. } e \notin E^{(i)}_{j_1}, e \notin E^{(i)}_{j_2}$ and $e \in M^{(i)}(V^{(i)}_{j_1}\cap V^{(i)}_{j_2})$, then we can use automatic rules $R_D$ to infer the ``context'' of $j_1$: $c^{(j_1)}_{j_2} := R_D(V^{(i)}_{j_1},V^{(i)}_{j_2}) \text{ s.t. } c^{(j_1)}_{j_2} \subseteq V^{(i)}_{j_2}$ and $M^{(i)}(V_{j_1}^{(i)} \sqcup c^{(j_1)}_{j_2})$ is connected. The same rule is applied in reverse to obtain $c^{(j_2)}_{j_1}$. The descriptions and examples for dataset-specific rules are given in Appendix~\ref{app:A2}.



There are various automated methods to obtain such  a fragmentation (e.g., \cite{fragmenter,brics, hgraph2graph}); 
some are integrated in the commonly used RDKit package \cite{rdkit}.
%
Nevertheless, we found that 
complex molecular datasets often benefit from fragmentations and rules tailored to the application domain, in the sense that they may better capture known domain knowledge and provide a strong structural prior. For this reason, we also designed and executed feasible, practical workflows
for annotating molecules and extracting the motifs.

\paragraph{Motif Graph Construction.}







Given a set of motifs, $V$, we describe our hierarchical abstraction over $V$. 
$G=(V,E)$ is a directed multigraph. Each $v \in V$ contains  both the motif 
graph $g_v$ and $\{v_{r_l}\}$, denoting the  possible “contexts” for attaching $g_v$ to another motif; 
that is, $\forall l, v_{r_l} \subseteq N(g_v)$, with $N(g)$ denoting the set of atom nodes of graph $g$. $v_R := \cup_l v_{r_l}$, and $\emptyset \neq v_B := N(g_v) \setminus v_R.$ Denoting $\sim$ to be the isomorphism relation, we construct $E$ by matching every pair of motifs $u, v$ and their contexts $(l_1, l_2)$ by finding corresponding subgraphs in $u_{B}$ and $v_{B}$ to match $u_{r_{l_1}}$ and $v_{r_{l_2}}$, as shown in Fig.~\ref{fig:methodG}~(c). Specifically, $(u, v, e_{l_1,l_2}) \in E \iff \exists b_2 \subseteq u_B, b_1 \subseteq v_B$ such that: 
\begin{align}
\label{eq:criteria-1}
    & g_u(u_{r_{l_1}}) \sim g_v(b_2) \\ \label{eq:criteria-2}
    & g_v(v_{r_{l_2}}) \sim g_u(b_1) \\ \label{eq:criteria-3}
    & g_{u}(u_{r_{l_1}} \cup b_1) \sim g_{v}(b_2 \cup v_{r_{l_2}})\\ \label{eq:criteria-4}
    & g_{u}(u_{r_{l_1}} \cup b_1) \text{ is connected }
\end{align}

$e_{l_1,l_2}$ is attributed with $u_{r_{l_1}}, v_{r_{l_2}}, b_1, b_2.$ 

The construction of the motif graph $G$ is in practice very efficient. For example, for the datasets we study, it is done under a minute when parallelized across $100$ CPU cores.
Details are given in Appendix~\ref{app:C}.

\subsection{The Molecular Design Space as Derivations of a Context-Sensitive Grammar Over Motif Graph} \label{sec-method-1}
We now define our context-sensitive grammar over $G$. We use the notations defined in the previous section to enumerate the set of production rules, $\mathcal{P}$, in our grammar. There is one initial rule $p_v \in \mathcal{P}$ for each motif $v$ in $G$, where the $LHS$ is $\mathcal{X}$, and the $RHS$ is the molecular graph $g_v$ with $u_B$ being the base atoms and $\{(u_{r_l})\}$ being the red atom sets that become ``options" for attachment. Then, there is exactly one production rule $p_{u,v,l_1,l_2}\in \mathcal{P}$ for each edge $(u, v, e_{l_1, l_2}) \in G$. This edge was attributed with $(u_{r_{l_1}}, v_{r_{l_2}}, b_1, b_2)$ during the construction of $G$. The application of the production rule then equates to attaching the fragment of $v$ to the fragment of $u$, at the attachment options keyed with $l_1, l_2$. In the language of graph grammars, the context of this production rule is hence the molecular graph $g_u(u_B \cup u_{r_{l_1}})$, with the requirement that the matched atoms for $u_{r_{l_1}}$ are red. Applying this production rule replaces the matched atoms for $u_{r_{l_1}}$ within the $LHS$ by $g_v(N(g_v) \setminus v_{r_{l_2}})$, where the red atom sets $\{v_{r_l} \mid v_{r_l} \cap v_{r_{l_2}} \neq \emptyset \}$ in $v$ are introduced as new options for attachment in the $RHS$. The random walk characterization arises out of the fact that if the $LHS$ molecule contains the context $g_u(u_B \cup u_{r_{l_1}})$, \textit{any} edge $(u, v, e_{l_1, l_2}) \in E$ can be traversed, possibly including self-loops and parallel edges since $G$ is a directed multidigraph.

\subsection{Molecules as Random Walks in the Design Space}

Intuitively, our representation of a molecule $M$ captures a derivation in the above-defined context-sensitive grammar. While prior work has modeled such derivations in large and complex tree structures (e.g., with auxiliary nodes for partial derivations) \cite{deg,ours}, we model it compactly in terms of a random walk over
the bidirectionally connected subgraph $H_M=(V_M, E_M)$ of $G$ given by the fragmentation of $M$\footnote{Refer to Appendix~\ref{app:B3} for how and why we augment $G$ with duplicates of the same motif.}; see Fig.~\ref{fig:methodG}~(a).
%
Observe that $G$ is a strong prior for constraining the design space and sufficient for describing the molecular structure of $M$, but $H_M$ misses the global distribution of which it is a sample of. 

\begin{figure*}[t]
  \centering
  \includegraphics[width=0.95\linewidth]{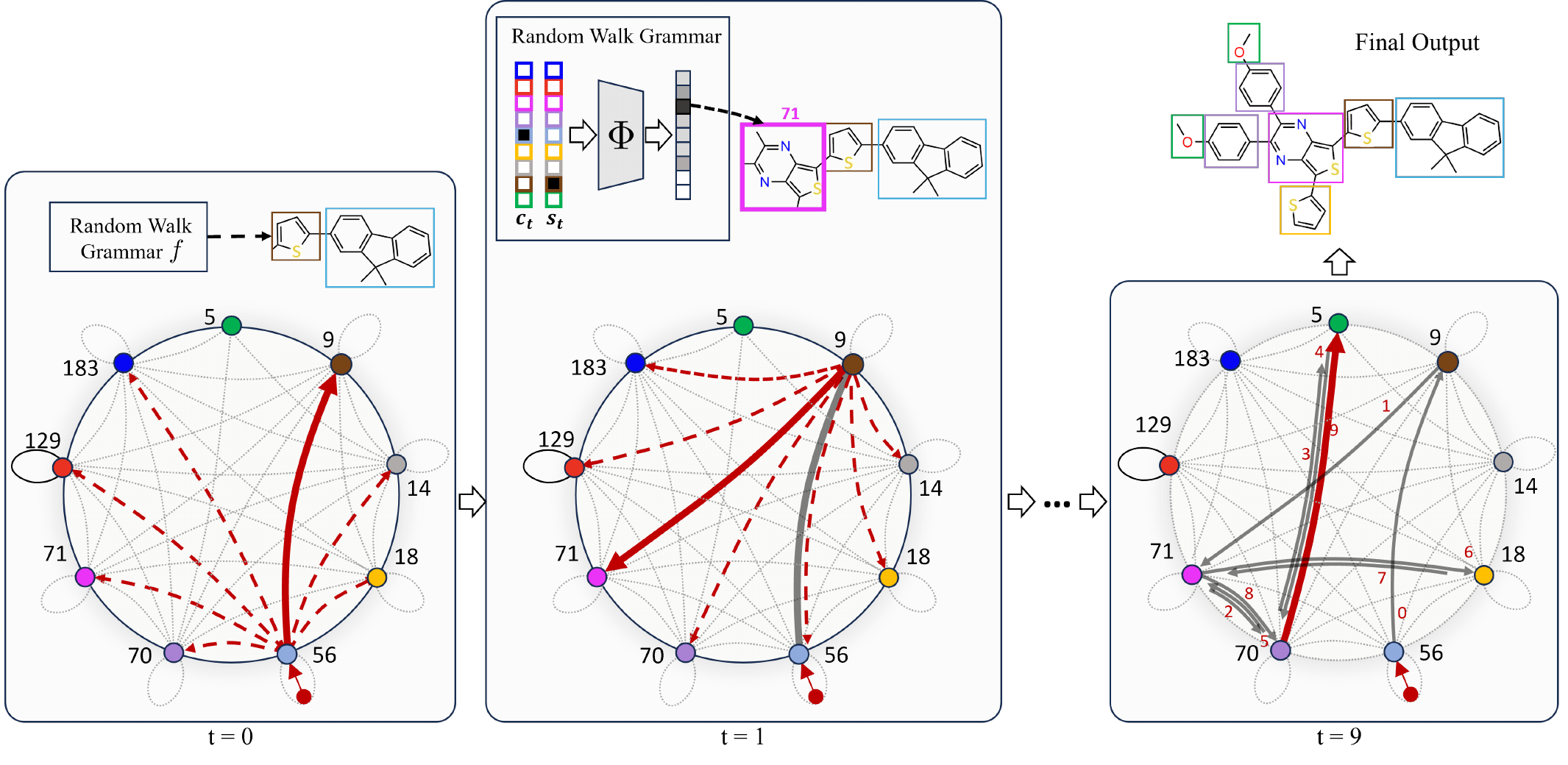}
  \caption{Illustration of our generation procedure: (t=1) our learnable grammar  parameterized by $\Phi$ samples a state transition $56\rightarrow 9$; (t=2) with the memory of having visited $\{56\}$, our grammar samples a state transition $\rightarrow 71$; (t=10) (bottom) our grammar samples a final transition $5$, which determines the molecular structure (top); our program's notation is $56\rightarrow 9\rightarrow 71 [\rightarrow 70\rightarrow 5]\rightarrow 70:1\rightarrow 5:1$}
  \label{fig:method2}
\end{figure*}

Our learnable component models this distribution and, at the same time, captures the features that characterize a specific molecule in terms of a random walk. 
More specifically, our final molecule representation is a directed-acyclic multi-graph
$\hat{H}_M =(V_M, \hat{E}_{M}, w_M)$ that linearizes $H_M$ into a random walk such that (1)~$\hat{E}_M \subseteq E_M$, (2)~$\hat{H}_M$ remains connected, and (3) there is an Euler path\footnote{In the case of monomers, the Euler path needs to be closed as monomers have the property of self-loops.}. (i.e., each edge is used exactly once) $v_0, v_1, \ldots, v_\ell$ over $(V_M, \hat{E}_M)$ with  $\hat{E}_M:=
\cup_i\{(v_i,v_{i+1})\}$; 
this path can be generated via a pre-order traversal that adds a reversed duplicate of the sub-trajectory when the stack contracts.
The last component, $w_M$ is the sequence $w_M:=p_0,p_1,\dots,p_{\ell-1}$ of probabilities given by the random walk; that is, $p_i$ represents the probability with which the edge between $v_i$ and $v_{i+1}$ was traversed in the presence of all nodes visited thus far, as shown in Fig.~\ref{fig:method2}. $w_M$ is parameterized by our learnable grammar, and $\hat{H}_M$ is explainable as a random walk of context-sensitive grammar rule applications.




\subsection{Learning Motif-based, Context-sensitive Grammars}

\subsubsection{Parameter Estimation}

For parameter estimation, we formulate the process of random walk as a graph heat diffusion process,
\begin{equation}
    \frac{\mathrm{d}x_t}{\mathrm{d}t} = L(\Phi, t) x_t,
\end{equation}
where $x_t\in \mathbb{R}^{|V|}$ represents the probability distribution of sampling motifs and $L(\Phi, t)\in \mathbb{R}^{|V|\times|V|}$ is a time-dependent graph Laplacian parameterized by $\Phi$. Here the initial condition of the diffusion process $x_0$ is a one-hot vector with the root of $\hat{H}_M$ as one. At every time step, the ground-truth $x_t$ follows the transition state of a random walk. 
In our implementation,  $L(\Phi, t)$ is calculated as
\begin{equation}
    L(\Phi, t) = D - \hat{W}(t), \hat{W}(t) = W + h(c_t;\phi)
\end{equation}
where $D\in \mathbb{R}^{|V|\times|V|}$ is the in-degree matrix of $G$, $h(\cdot;\phi)$ is a memory-sensitive adjustment layer, and $c_t$ is a set-based memory of all nodes visited thus far. If $p^{(t)}$ is the current state of the random walk, the set-based memory, $c^{({t+1})}$, is updated as follows: $c^{(t+1)} \leftarrow \frac{t}{t+1} \cdot c^{(t)}+\frac{1}{t+1}\cdot p^{(t)}$. This set-based memory mechanism has precedents in graph theory literature. The learnable parameters are $\Phi = (W, \phi)$. Further motivation of the set-based memory mechanism is in Appendix \ref{app:C3} and the full training algorithm can be found in Appendix \ref{app:D}.

\subsubsection{Training for a Downstream Task}
\textbf{Property Prediction} Our grammar-induced molecular graph representation $\hat{H}_i$ allows for applying an off-the-shelf graph neural network $\mathcal{F}_{\Theta}$ to solve a given prediction task; in our evaluation, we used GIN \cite{gin}. Given a property value $y^{(i)} \in \mathbb{R}$ with each molecule $M^{(i)}$, we apply a linear head $f_{\theta}$ and application-specific loss function $\mathcal{L}$ (e.g., MSE for regression or cross-entropy for classification).

\textbf{End to End Training} Our grammar-based representation can further be optimized via end-to-end training of $\Phi$. Typically, we first train $\Phi$ to convergence under our MC-based objective, then train $\Theta$ to convergence under eq~\ref{eq:loss} on the representations induced by $\Phi$. Finally, we freeze $\Theta$ and finetune $\Phi$ to convergence. Alternatively, we train $\Phi$ and $\Theta$ together, end to end, by using the following differentiable objective,
%
\begin{align}
\label{eq:loss}
\Tilde{\mathcal{L}}(D ; \Theta, \theta, \Phi) 
&= \mathbb{E}_{\hat{H}_M(\cdot;\Phi)}[ \mathcal{L}(f_{\theta}(\mathcal{F}_{\Theta}(\hat{H}_M, y)] \\
&= \frac{1}{|D|}\sum_{i=1}^{|D|} \mathcal{L}(f_{\theta}(\mathcal{F}_{\Theta}(\hat{H}_M^{(i)})), y^{(i)}),
\end{align}
where we estimate the expectation using the training samples from training dataset $D$.

\subsubsection{Molecular Generation} To generate a molecule $M$, we apply the learned grammar forward to sample edges to traverse during the random walk, as depicted in Fig.~\ref{fig:method2}. Each sampled edge either attaches a new motif to the current $M$, or backtracks to a previous motif. Details on the algorithm are given in Appendix~\ref{app:F}.




%% file: body/4.results.tex
\section{Results \& Analysis}

Our experiments quantitatively answer the following questions: 1) How well does our method perform on property prediction for our setting of interest? 2) How well does our representation work for the generation of novel molecules, compared with both SOTA symbolic and deep molecular generative models? We also include three ablations to answer: 3) How important are expert motifs, compared to heuristic-based motifs? 4) How data-efficient and runtime-efficient is our method? 5) How does our method compare with other motif-based predictors? Our qualitative analysis answers the following questions: 6) How interpretable is our learnt grammar to an expert? 7) How can we analyze the model’s learnt representations?


\begin{figure}[b]
  \centering 
  \includegraphics[width=0.5\textwidth]{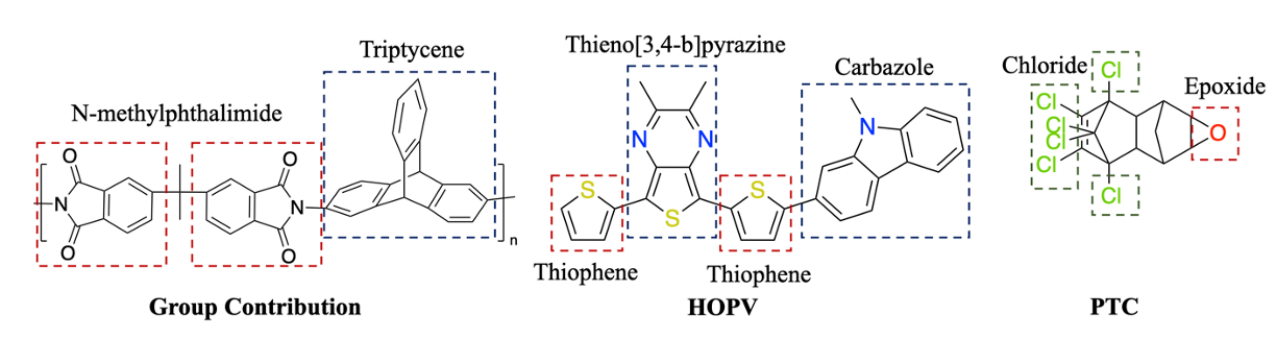}
  \caption{Example molecules from GC, HOPV, and PTC. These datasets are characterized by modular substructures that correspond to meaningful chemical functional groups.
  }
  \label{examples}
\end{figure}

\subsection{Datasets and Baselines} 

\begin{table}[h!]
\label{tab:data}
\centering
\caption{Average size of our hierarchical representation $H_M$ over each dataset, with expert vs heuristic motifs.}
\resizebox{\columnwidth}{!}{%
\begin{tabular}{|c|cc|cc|cc|}
\toprule
\textbf{Dataset}         & \multicolumn{2}{c|}{\textbf{GC}}                       & \multicolumn{2}{c|}{\textbf{HOPV}}                     & \multicolumn{2}{c|}{\textbf{PTC}}                      \\ \midrule
\textbf{Expert}          & \multicolumn{1}{c|}{Yes}             & No              & \multicolumn{1}{c|}{Yes}             & No              & \multicolumn{1}{c|}{Yes}             & No              \\ \midrule
\textbf{Avg. $|$H\_M$|$} & \multicolumn{1}{l|}{$7.3 \pm 2.8$} & $3.8 \pm 2.2$ & \multicolumn{1}{l|}{$5.4 \pm 1.9$} & $6.5 \pm 2.9$ & \multicolumn{1}{l|}{$3.6 \pm 2.4$} & $2.1 \pm 1.4$ \\ \midrule
\end{tabular}
}

\end{table}

\vspace{-1.5ex}
\begin{figure}[h]
  \centering
  \includegraphics[width=0.5\textwidth]{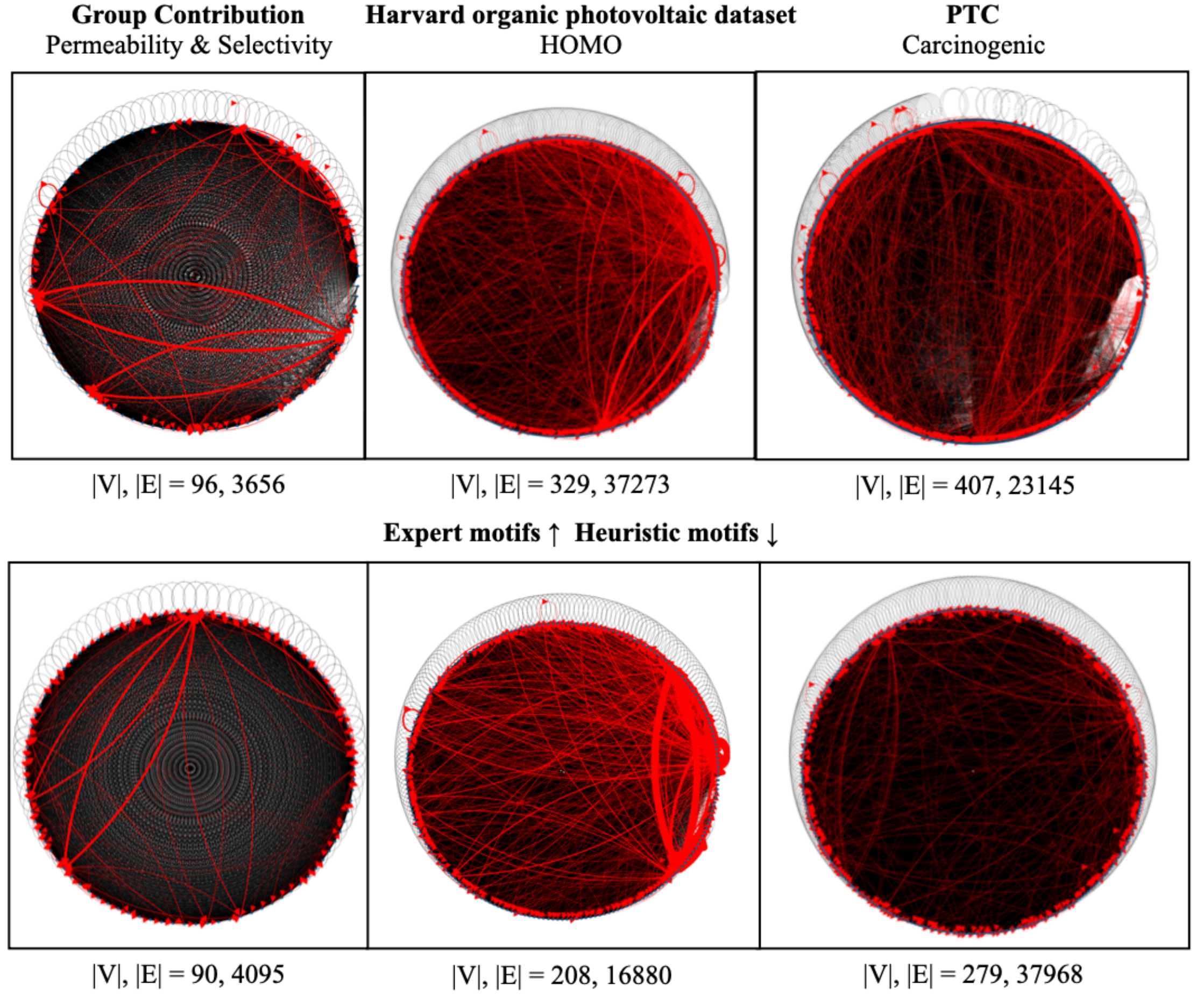}
  \caption{Visualization of our motif graph $G$; black edges indicate matched motif pairs, thickness of \textcolor{red}{red} edges correspond to the numbers of $H_M$ that traverse it.}
  \label{fig:data-stats}
  \vspace{-3ex}
\end{figure}

\textbf{Group Contribution (GC)} \cite{microporous, group-contrib-1, group-contrib-2}\textbf{.}
114 molecules, characterized in terms of gas separation.
Their functional groups contribute significantly to maintaining the structures and properties of 3D scaffold building blocks in polymer self-assembly, which in turn play a significant role in gas separation processes, 
important in gas and oil industry. We used existing monomers \cite{microporous} and compilations of groups \cite{group-contrib-1, group-contrib-2} for inferring the fragmentations.

\textbf{The Harvard organic photovoltaic dataset (HOPV)} \cite{hopv}\textbf{.}
316 molecules, 
applied to aid in the design of 
organic solar cells, 
%
with detailed information pertinent to organic photovoltaic performance metrics. 
The molecules contain motifs which are among the most significant functional groups for conducting/electroactive materials \cite{Swager17} and photovoltaic properties \cite{Yuan22}. 
We extracted motifs important for high HOMO values and enhanced electron delocalization, 
which are critical for photovoltaic efficiency; see Appendix~\ref{app:G} for details. 

\textbf{Predictive Toxicology Challenge (PTC)} \cite{Helma01}\textbf{.}
344 small chemical compounds characterized by very distinct functional groups known for their carcinogenic properties or liver toxicity \cite{Miller49}, with reported values for rats. 
We specifically segmented it into functional groups that majorly contribute to the improvement of compounds’ toxicity \cite{Hughes15}. Examples from each dataset are shown in Figure~\ref{examples}. \\\\
\textbf{Baselines.}
To address question 1), we compare with pretrained GNNs (PN \cite{PN} and Pre-trained GIN \cite{pretrainGIN}), a specialized GNN model for property prediction (wD-MPNN \cite{aldeghi2022graph}), two SOTA pretrained models for molecular representation learning (MolCLR \cite{molCLR} and UniMol \cite{unimol}) and two SOTA subgraph-based methods (ESAN \cite{esan} and HM-GNN \cite{hmgnn}). To address question 2), we compare with both Geo-DEG, the SOTA on small dataset property prediction, and its generative variant, DEG, for molecular generation.
\subsection{Results}
We report the mean absolute error (MAE) and coefficient of determination ($R^2$) over normalized prediction values for GC and HOPV, and the accuracy and AUC for PTC. For each (dataset, property) pair, we perform an 80-20 train-test split over 3 random seeds and report the mean and standard deviation. For molecular generation, we report commonly used metrics \cite{Moses, deg}:
Validity/Uniqueness/Novelty: Percentage of chemically valid/unique/novel molecules;  
Diversity: Average pairwise molecular distance among generated molecules;  Retro* Score (RS): Success rate of Retro* \cite{retro-star} which was trained to find a retrosynthesis path to build a molecule from a list of commercially available ones. We add the metric of Membership, which tests whether certain motif(s) characteristic of membership to the chemical class are present, primarily as a sanity check. Our method, by design, can achieve 100\% if the random walk initializes at the characteristic motif. See \ref{app:A1} for further discussion.
\subsubsection{Property prediction}

To answer question 1), we see in Table \ref{table:results1} that our method, with expert motifs, achieves the best $R^2$ by a wide margin of $0.10$ and $0.06$ over the second best method on regression datasets GC and HOPV and the highest accuracy on PTC. With heuristic motifs, our method remains competitive to Geo-DEG, achieving higher $R^2$ on both regression datasets and accuracy within standard deviation on PTC. Interestingly, using heuristic-based motifs in HOPV, achieves significantly (27\%) lower MAE than expert-based motifs and Geo-DEG. To answer question 3), we see that the ablation suggests expert motifs are generally better, but may be more sensitive to outliers than heuristic-based motifs. We observe experts are generally better at identifying special cases that heuristics are unaware of, but heuristics are more consistent. This reflects how $R^2$ is generally more sensitive to outliers than MAE. We describe our experts’ annotation criteria in Appendix \ref{app:A} and do an in-depth case study on HOPV in Appendix \ref{app:G}.

\begin{table*}[h!]
\centering
\tiny
\caption{Results on property prediction (best result \textbf{bolded}, second-best \underline{underlined}). The datasets we include have expert-annotated motifs. We also report Ours (w/o expert) as an ablation without expert motifs.}
\label{table:results1}
\resizebox{2.1\columnwidth}{!}{%
\begin{tabular}{|@{}>{\centering\arraybackslash}p{1cm}>
{\centering\arraybackslash}p{0.7cm} |C{1.2cm}|C{1.4cm}|C{1.2cm}|C{1.6cm}|C{1.2cm}|C{1.2cm}|C{1.2cm}|C{1.2cm}|C{1.2cm}|C{1.2cm}|C{1.2cm}|}

\toprule
\multicolumn{2}{c|}{\multirow{2}{*}{\backslashbox{Datasets}{Methods}}} & \multirow{2}{*}{wD-MPNN} & \multirow{2}{*}{ESAN} & \multirow{2}{*}{HM-GNN} & \multirow{2}{*}{\shortstack{PN\\(finetuned)}} & \multirow{2}{*}{\shortstack{Pre-trained GIN\\(finetuned)}} & \multirow{2}{*}{MolCLR} & \multirow{2}{*}{Unimol} & \multirow{2}{*}{Geo-DEG}  & \multirow{2}{*}{\textbf{Ours}} & \multirow{2}{*}{\shortstack{\textbf{Ours}\\(w/o expert)}}\\
& & & & & & & & & &\\
\midrule
\multirow{2}{*}{\textbf{Group}} & \textbf{MAE} $\downarrow$ & $0.47 \pm 0.09$ & $0.51 \pm 0.06$ & $0.34 \pm 0.12$ & $0.76 \pm 0.30$ & $0.68 \pm 0.05$ & \underline{$0.26 \pm 0.10$} & $0.38 \pm 0.13$ & \underline{$0.26 \pm 0.11$}  & $\mathbf{0.25 \pm 0.09}$ & $0.27 \pm 0.08$\\
& $\mathbf{R^2} \uparrow$ & $0.41 \pm 0.12$ & $-0.39 \pm 0.62$ & $0.56 \pm 0.20$ & $-7.56 \pm -7.71$ & $0.19 \pm 0.09$ & $0.68 \pm 0.20$ & $0.47 \pm 0.25$ & $0.70 \pm 0.20$ & $\mathbf{0.80 \pm 0.15}$ & \underline{$0.74 \pm 0.15$}\\
\midrule
\multirow{2}{*}{\textbf{HOPV}} & \textbf{MAE} $\downarrow$ & $0.36 \pm 0.03$ & $0.37 \pm 0.02$ & $0.40 \pm 0.02$ & $0.42 \pm 0.02$ & $0.38 \pm 0.02$ & $0.34 \pm 0.03$ & $0.31 \pm 0.03$ & \underline{$0.30 \pm 0.02$} & \underline{$0.30 \pm 0.05$} & $\mathbf{0.22 \pm 0.15}$\\
& $\mathbf{R^2} \uparrow$ & $0.69 \pm 0.04$ & $0.66 \pm 0.06$ & $0.65 \pm 0.05$ & $0.65 \pm 0.04$ & $0.66 \pm 0.03$ & $0.68 \pm 0.03$ &  $0.70 \pm 0.02$ & \underline{$0.74 \pm 0.03$} & $\mathbf{0.80 \pm 0.06}$ & \underline{$0.77 \pm 0.12$}\\
\midrule
\multirow{2}{*}{\textbf{PTC}} & \textbf{Acc} $\uparrow$ & $0.67 \pm 0.06$ & $0.64 \pm 0.08$ & $0.66 \pm 0.07$ & $0.61 \pm 0.08$ & $0.62 \pm 0.09$ & $0.60 \pm 0.03$ & $0.57 \pm 0.05$ & \underline{$0.69 \pm 0.07$} & $\mathbf{0.70 \pm 0.01}$ & $0.67 \pm 0.02$\\
& \textbf{AUC} $\uparrow$ & \underline{$0.70 \pm 0.05$} & $0.68 \pm 0.06$ & $0.69 \pm 0.06$ & $0.65 \pm 0.07$ & $0.66 \pm 0.07$ & $0.66 \pm 0.05$ & $0.67 \pm 0.06$ & $\mathbf{0.71 \pm 0.07}$ & $0.69 \pm 0.03$ & $0.66 \pm 0.05$\\
\bottomrule
\end{tabular}
}
\vspace{-2.5ex}
\end{table*}

\subsubsection{Molecular generation}\label{sec:generation}

To answer question 2), we see in \cref{molgen} that our method produces comparably more diverse molecules than the training dataset (+0.03 on HOPV, -0.01 on PTC) and significantly more synthesizable molecules than the previous SOTA, DEG (+39\% on HOPV, +22\% on PTC). On HOPV, our retrosynthetic planner finds synthesis paths at a 14\% \textit{higher} rate on our novel molecules than the \textit{original} dataset, a careful collation of experimental photovoltaic designs \cite{hopv}. This is encouraging to experimentalists whose work is contingent on the designs' feasibility for synthesis. We also compare our methods with established VAE-based molecular generative models such as \cite{jtvae} and its follow-up work \cite{hgraph2graph} which includes larger structural motifs. Furthermore, we modified the implementation of Hier-VAE to incorporate our epert motifs. For all three cases, we follow the default settings, train until convergence, and use the checkpoint with the lowest loss to sample 1000 molecules. We observe that both VAE-based methods struggle to generate sufficiently unique molecules, with only 11\%-43\% (HOPV) and 8\%-28\% (PTC) of the 1000 generated molecules being unique. This is despite sampling from a Gaussian noise distribution. Meanwhile, our model generates 100\% unique and novel molecules, while ensuring a high internal diversity second only to DEG. For reference, \cite{jtvae, hgraph2graph} trained and evaluated their methods on ~250K molecules extracted from ZINC \cite{zinc} and a polymer dataset containing 86K polymers. Meanwhile, our datasets contain only ~100-300 molecules and, in the case of HOPV, feature much larger molecules. Rather than using an encoder-decoder setup which requires significantly more data to learn the mapping to and from a latent space, our generative model explicitly captures the transition probabilities over traversing the symbolic space of structural motifs. Our grammar derivation can easily be conditioned by a set-based memory to apply a diverse set of transition rules. This leads to more unique, diverse, and, most importantly, synthesizable structures.

\begin{table}[h!]
\centering
\vspace{-3ex}
\caption{Results on molecular generation for HOPV (top) and PTC (bottom); for both datasets, we generate 1000 novel molecules. Refer to Appendix \ref{app:A1} for more details on Membership.}
\label{molgen}
\small

\resizebox{\columnwidth}{!}{%
\begin{tabular}{|c|c|cccccc|}

\toprule
Datasets & Methods            & Valid & Unique & Novel & Diversity & RS   & Memb. \\ \midrule

\multirow{2}{*}{HOPV}     & Train Data         & 100\% & 100\%  & N/A   & 0.86      & 51\% & 100\% \\ \cmidrule{2-8}
         & DEG                & 100\% & 98\%   & 99\%  & {0.93}      & 19\% & 46\%  \\
         & JT-VAE             & 100\% & 11\%   & 100\% & 0.77      & {99\%} & 84\%  \\
         & Hier-VAE           & 100\% & 43\%   & 96\%  & 0.87      & 79\% & 76\%  \\
         & Hier-VAE (+expert) & 100\% & 29\%   & 92\%  & 0.86      & 84\% & 82\%  \\
         & Ours               & 100\% & 100\%  & 100\% & 0.89      & 58\% & 71\%  \\ \midrule
\multirow{2}{*}{PTC}      & Train Data         & 100\% & 100\%  & N/A   & 0.94      & 87\% & 30\%  \\ \cmidrule{2-8}
         & DEG                & 100\% & 88\%   & 87\%  & {0.95}      & 38\% & 27\%  \\
         & JT-VAE             & 100\% & 8\%    & 80\%  & 0.83      & {96\%} & 27\%  \\
         & Hier-VAE           & 100\% & 20\%   & 85\%  & 0.91      & 92\% & 25\%  \\
         & Hier-VAE (+expert) & 100\% & 28\%   & 75\%  & 0.93      & 90\% & 17\%  \\
         & Ours               & 100\% & 100\%  & 100\% & 0.93      & 60\% & 22\% \\ \bottomrule

\end{tabular}
}
\vspace{-3ex}
\end{table}

%% file: body/5.discussion_analysis.tex
\subsection{Ablations}

\subsubsection{Ablation: vary training dataset size}

\begin{figure}[h]
  \centering
  \includegraphics[width=0.5\textwidth]{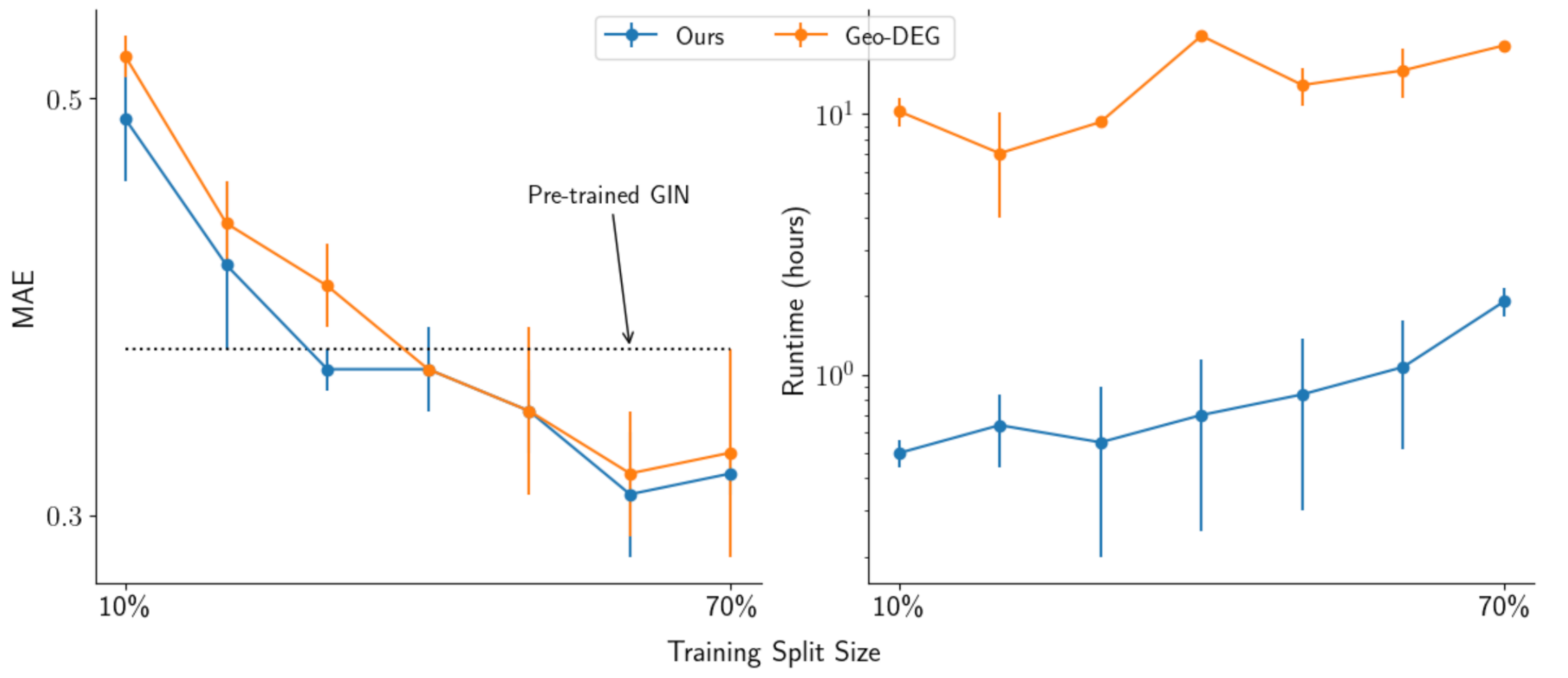}
  \caption{
  Varying the training dataset size from 10-70\%.}\label{size-ablation}
  \vspace{-2.5ex}  
\end{figure}

To answer question 4), we conduct an ablation study in \cref{size-ablation} over the training split size to study how data and runtime-efficient our method is in comparison with Geo-DEG. Our method performs strictly better on MAE as the training set is reduced from 70\% to 10\%. This is in addition to the method running an order of magnitude faster, highlighting gains in both data efficiency and runtime efficiency.



\begin{table*}[h]
\small
\centering
\scriptsize
\caption{Ablation study on overfitting and generalization, vs other motif-based baselines, with and w/o expert motifs. Best result is \textbf{bolded}.}
\label{bag-of-motifs}
\resizebox{2.1\columnwidth}{!}{%
    \begin{tabular}{|C{2.4cm}|C{0.75cm}|C{0.75cm}|C{0.75cm}|C{0.75cm}|C{0.75cm}|C{0.75cm}|C{0.75cm}|C{0.75cm}|C{0.75cm}|C{0.75cm}|C{0.75cm}|C{0.75cm}|} \toprule
         Ablation/Dataset & \multicolumn{4}{c|}{HOPV} & \multicolumn{4}{c|}{PTC} & \multicolumn{4}{c|}{Group Contribution}\\
         \midrule
         & Train MAE $\downarrow$ & Train $R^2$ $\uparrow$ & Test MAE $\downarrow$ & Test $R^2$ $\uparrow$ & Train Acc $\uparrow$ & Train AUC $\uparrow$ & Test Acc $\uparrow$ & Test AUC $\uparrow$ & Train MAE $\downarrow$ & Train $R^2$ $\uparrow$ & Test MAE $\downarrow$ & Test $R^2$ $\uparrow$\\
         \midrule
         Bag-of-Motifs & \shortstack{$0.014 \pm$\\$ 0.002$} & \shortstack{$0.997 \pm$\\$ 0.001$} & \shortstack{${0.486 \pm}$\\${ 0.025}$} & \shortstack{${0.489 \pm}$\\${ 0.062}$} & \shortstack{${0.996 \pm}$\\${ 0.000}$} & \shortstack{$\mathbf{1.000\pm}$\\$\mathbf{ 0.000}$} & \shortstack{$0.529 \pm$\\$ 0.031$} & \shortstack{$0.609 \pm$\\$ 0.031$} & \shortstack{$\mathbf{0.000 \pm}$\\$\mathbf{ 0.000}$} & \shortstack{$\mathbf{1.000\pm}$\\$\mathbf{ 0.000}$} & \shortstack{${0.481\pm}$\\${ 0.174}$} & \shortstack{${0.257\pm}$\\${ 0.453}$} \\         
         \midrule
         Bag-of-Motifs (+expert) & \shortstack{$\mathbf{0.011 \pm}$\\$\mathbf{ 0.004}$} & \shortstack{$\mathbf{1.000 \pm}$\\$\mathbf{ 0.000}$} & \shortstack{$0.521 \pm$ \\$ 0.031$} & \shortstack{$0.446 \pm$\\$ 0.125$} & \shortstack{${0.996 \pm}$\\${ 0.000}$} & \shortstack{$\mathbf{1.000 \pm}$\\$\mathbf{ 0.000}$} & \shortstack{${0.581 \pm}$\\${ 0.018}$} & \shortstack{${0.612 \pm}$\\${ 0.029}$} & \shortstack{$\mathbf{0.000 \pm}$\\$\mathbf{ 0.000}$} & \shortstack{$\mathbf{1.000 \pm}$\\$\mathbf{ 0.000}$} & \shortstack{$0.493 \pm$\\$ 0.143$} & \shortstack{$0.214 \pm$\\$ 0.404$}\\      
         \midrule
         HM-GNN & \shortstack{${0.366 \pm}$\\${ 0.035}$} & \shortstack{${0.686 \pm}$\\${ 0.066}$} & \shortstack{$0.473 \pm$ \\$ 0.019$} & \shortstack{$0.441 \pm$\\$ 0.065$} & \shortstack{${0.915 \pm}$\\${ 0.033}$} & \shortstack{${0.966 \pm}$\\${ 0.016}$} & \shortstack{$\mathbf{0.710 \pm}$\\$\mathbf{ 0.023}$} & \shortstack{${{0.678 \pm}}$\\${{ 0.040}}$} & \shortstack{${0.281 \pm}$\\${ 0.064}$} & \shortstack{${0.717 \pm}$\\${ 0.137}$} & \shortstack{$0.362 \pm$\\$ 0.113$} & \shortstack{$0.592 \pm$\\$ 0.202$}\\      
         \midrule        
         HM-GNN (+expert) & \shortstack{${0.201 \pm}$\\${ 0.009}$} & \shortstack{${0.895 \pm}$\\${ 0.019}$} & \shortstack{$0.451 \pm$ \\$ 0.025$} & \shortstack{$0.408 \pm$\\$ 0.095$} & \shortstack{$\mathbf{0.999 \pm}$\\$\mathbf{ 0.002}$} & \shortstack{$\mathbf{1.000 \pm}$\\$\mathbf{ 0.000}$} & \shortstack{${0.681 \pm}$\\${ 0.024}$} & \shortstack{${0.587 \pm}$\\${ 0.075}$} & \shortstack{${0.185 \pm}$\\${ 0.016}$} & \shortstack{${0.926 \pm}$\\${ 0.039}$} & \shortstack{${0.345 \pm}$\\$ {0.149}$} & \shortstack{${0.547 \pm}$\\$ {0.295}$}\\   
         \midrule         
         Ours (-expert) & \shortstack{${0.075 \pm}$\\${ 0.003}$} & \shortstack{${0.990 \pm}$\\${ 0.001}$} & \shortstack{$\mathbf{0.288 \pm}$ \\$\mathbf{0.048}$} & \shortstack{${0.765 \pm}$\\${0.146}$} & \shortstack{${0.994 \pm}$\\${ 0.001}$} & \shortstack{${0.999 \pm}$\\${ 0.000}$} & \shortstack{${0.671 \pm}$\\${ 0.020}$} & \shortstack{${0.659 \pm}$\\${ 0.047}$} & \shortstack{${0.044 \pm}$\\${ 0.015}$} & \shortstack{${{0.995 \pm}}$\\${ 0.004}$} & \shortstack{${0.268 \pm}$\\$ {0.084}$} & \shortstack{${0.738 \pm}$\\$ {0.148}$}\\            
         \midrule         
         Ours & \shortstack{${0.045 \pm}$\\${ 0.003}$} & \shortstack{${0.996 \pm}$\\${ 0.001}$} & \shortstack{${0.295 \pm}$ \\${0.049}$} & \shortstack{$\mathbf{0.796 \pm}$\\$\mathbf{0.105}$} & \shortstack{${0.996 \pm}$\\${ 0.000}$} & \shortstack{$\mathbf{1.000 \pm}$\\$\mathbf{ 0.000}$} & \shortstack{${0.705 \pm}$\\${ 0.007}$} & \shortstack{$\mathbf{0.711 \pm}$\\$\mathbf{ 0.018}$} & \shortstack{${0.028 \pm}$\\${ 0.007}$} & \shortstack{${{0.998 \pm}}$\\${ 0.002}$} & \shortstack{$\mathbf{0.222 \pm}$\\$ \mathbf{0.079}$} & \shortstack{$\mathbf{0.819 \pm}$\\$ \mathbf{0.137}$}\\             
         \bottomrule
    \end{tabular}
}
\end{table*}

\subsubsection{Comparison with motif-based baselines}
To answer question 5), we compare with two baselines. The first, Bag-of-Motifs, ablates our hierarchical information and retains only the motif co-occurrence information. For each molecule, we obtain a feature vector that concatenates a) the occurrence counts of all motifs and b) the Morgan fingerprint of the molecule. 
We train an XGBoost regressor/classifier on top of these features (details in Appendix \ref{app:E}). As shown in Table \ref{bag-of-motifs}, this baseline has enough capacity to overfit the training data but fails to generalize. This allows us to conclude in the absence of a proper representation, motif occurrence information is not sufficient for generalization. Interestingly, expert-level motifs are not superior to heuristic-based motifs in this featurization. This suggests that the quality of motifs are not relevant in the absence of a hierarchical representation that incorporates the fine-grained features of each individual motif. The second, HM-GNN \cite{hmgnn}, is a SOTA motif-based property predictor that explicitly models motif-molecule and motif-motif relationships using a hetereogenous graph. Furthermore, we endowed the method with our expert motifs since the vanilla version only considers bonds and rings. On both regression datasets, HM-GNN avoids overfitting but does not catch up to our method's generalization capability. Endowing HM-GNN with our expert motifs enables better fitting of the training data but further hinders generalization. On PTC, HM-GNN is competitive with our method in accuracy but shows a discrepancy in terms of AUC. This is concerning as a lower AUC may imply higher sensitivity to class imbalance (in PTC, there are 45\% more negatives than positives) and classification thresholds. Meanwhile, our method can both 1) completely fit the training data ($>0.99$ $R^2$, $>99\%$ Acc/AUC), and 2) generalize to the test data, with further regularization potentially leading to even better results. We believe our motif-based representation carries better inductive biases, integrates better with expert motifs, and demonstrates stronger empirical performance.

\subsection{Analysis}
\subsubsection{Rule extraction from grammar} \label{analysis-rule-extraction}
To answer question 6), we extract context-sensitive grammar rules from our trained model. We perform best-first search over random walk trajectories, beginning with base trajectories corresponding to each group $v \in G$. We only expand trajectories with transition probabilities above a minimum threshold. Each trajectory that reaches a transition with probability of 1 is extracted as a “hard” context-sensitive rule. We depict two such rules in Figure~\ref{fig:rule-extraction}, with a more exhaustive compilation in Appendix \ref{app:F1}.

\begin{figure}[h]
  \centering
  \includegraphics[width=0.48\textwidth]{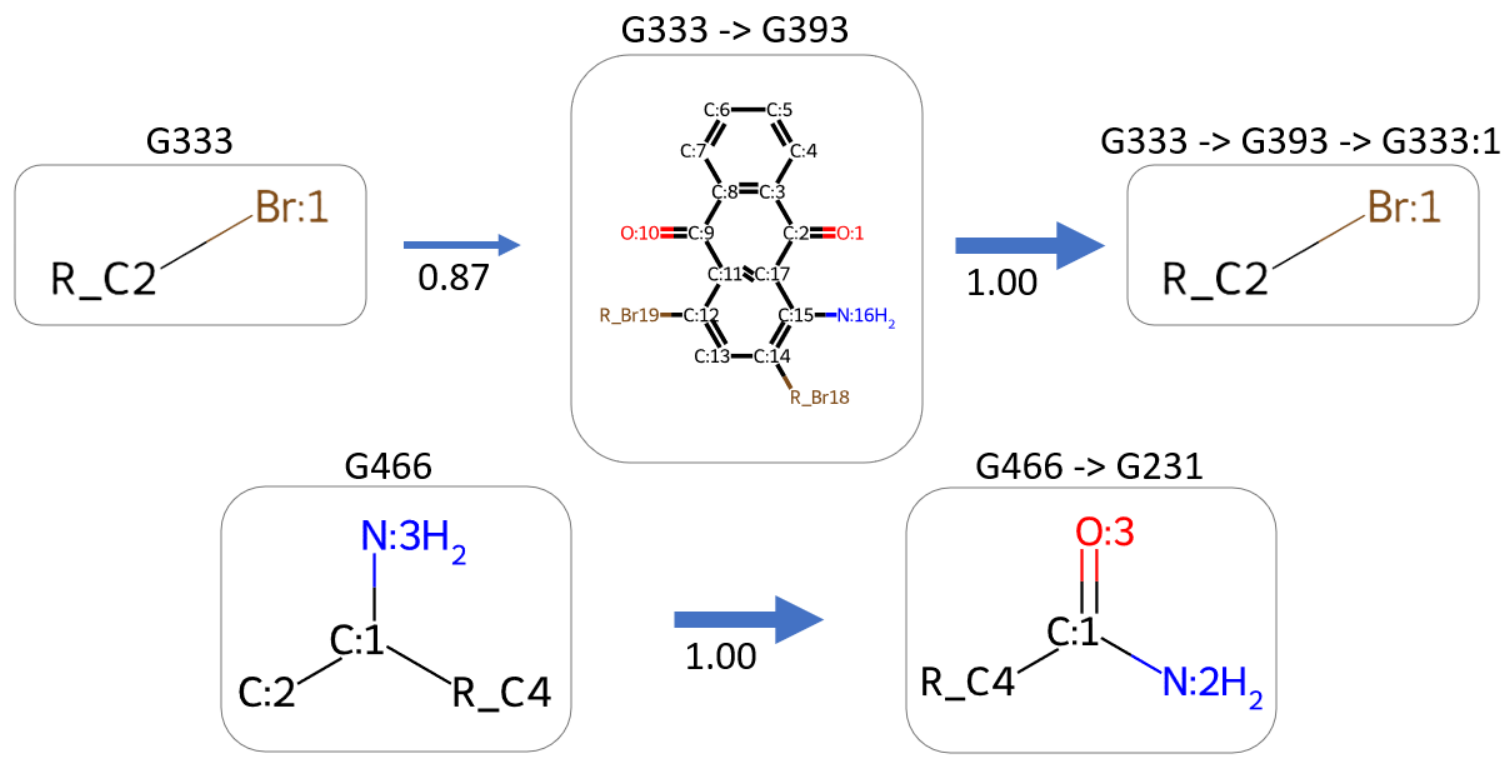}
  \caption{We visualize two hard context-sensitive rules on PTC that correspond to design principles of the addition of halogen groups to further improve molecular toxicity.}
  \label{fig:rule-extraction}
\end{figure}
\vspace{-1ex}

Figure~\ref{fig:rule-extraction} shows how our model recovers a set of design principles used to facilitate the synthesis of functional molecules and grounded in the structure-property relationship of PTC. Consider the transformation of the triple benzene derivative molecule (labeled as [‘G333’, ‘G393’]) with the addition of bromide moiety (labeled as [‘G333’, ‘G393’, ‘G333:1’]). In this instance, the central moiety, G393, is characterized by two symmetrical ketone groups and two bromides adjoined to the aromatic ring. This configuration markedly enhances the molecule’s toxicity. Moreover, by strategically positioning additional binding sites on the aromatic ring, the software augments the molecule with two extra bromide groups, G333, thereby exacerbating its hepatotoxicity. In another example, the molecule with an ammonia group (labeled as [‘G466’]) transforms with an additional ketone group (labeled as [‘G466’, ‘G231’]). Here, the presence of a C=O double bond within an acetamide group is a key contributor to hepatotoxicity. 

\subsubsection{Two-dimensional T-SNE on $H_M$ versus pretrained representations}\label{sec:tsne}

\begin{figure}[h]
  \centering
  \includegraphics[width=.45\textwidth]{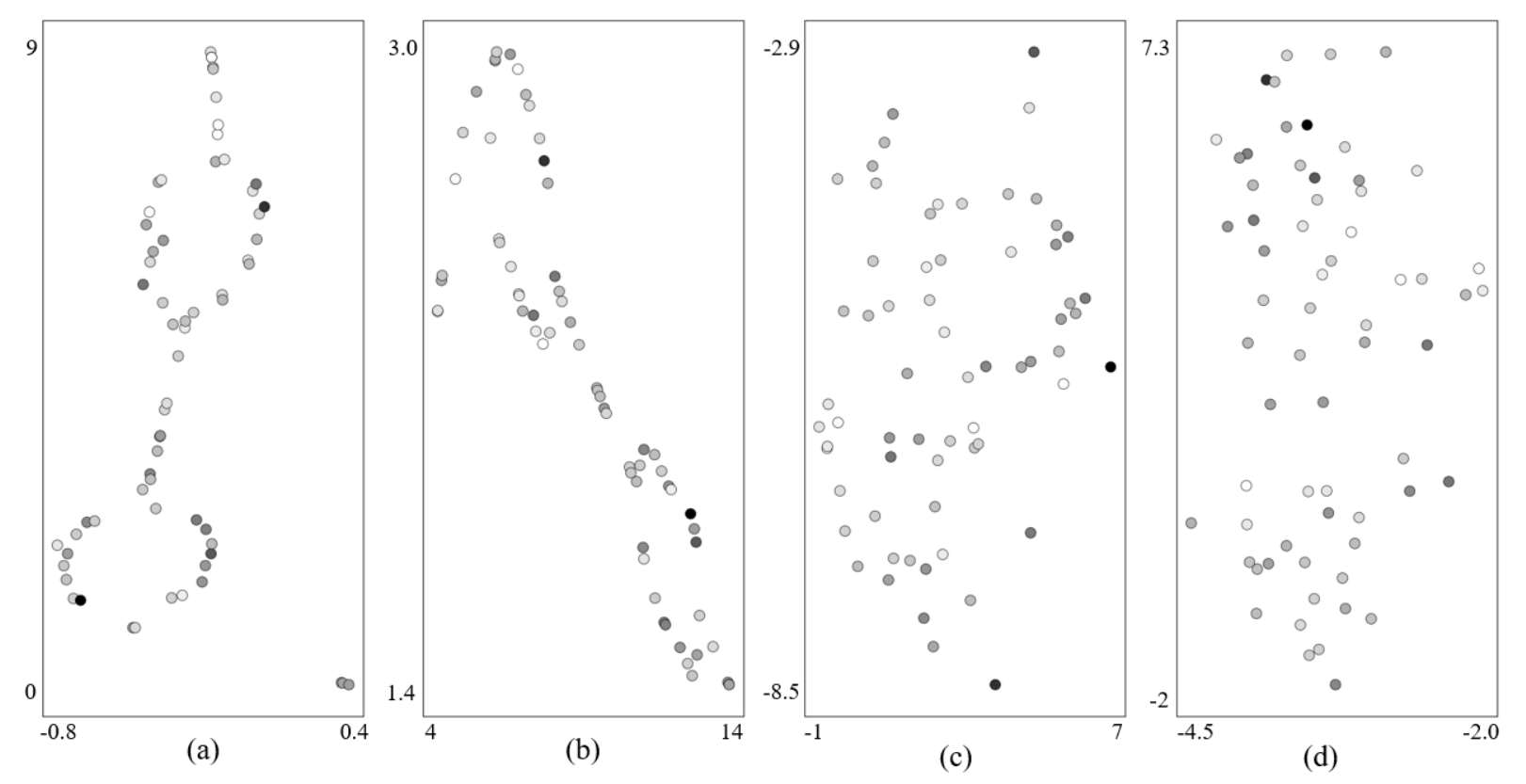}
  \caption{Final layer representations from: a) Our method b) Our method (-expert) c) Pre-trained GIN d) HM-GNN. We apply a grayscale coloring map using the normalized value of the desired property (the darker the dot, the higher the HOMO).}
  \label{tsne}
\end{figure}
\begin{figure}[h]
  \centering
  \includegraphics[width=0.45\textwidth]{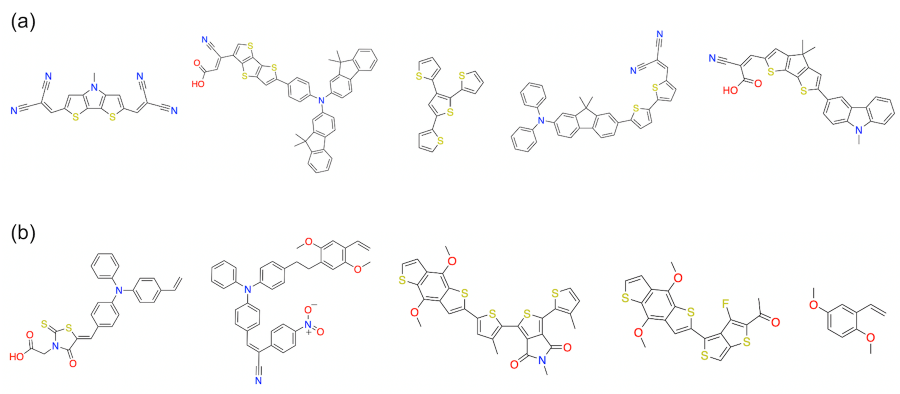}
  \caption{Examples of top HOMO value compounds with group (a) from the top cluster and group (b) from the bottom cluster.}
  \label{fig:Clusters}
  \vspace{-2.5ex}
\end{figure}

To answer question 7), we analyze the 2D t-SNE embeddings of various methods' final layer representations of 64 test set molecules on HOPV. As shown in Figure~\ref{tsne}, our method is unique in extracting visually meaningful representations. High HOMO molecules were identified from the visual clusters for structural analysis. Molecules in the upper cluster as illustrated in Figure~\ref{fig:Clusters} often have structures promoting electron delocalization, like carbonohydrazonoyl dicyanide, while those in the lower cluster have electron-donating groups or structures increasing steric hindrance to boost HOMO values as shown in Figure~\ref{fig:Clusters}. These two structural features correspond to the two primary ways to design molecules with high HOMO values. These findings aid the search for novel molecules with desirable photovoltaic properties. As 2D t-SNE is not a universal way to analyze representations, we also visualize the agreement between embedding similarity and structural similarity using a $64\times 64$ grid. This is can be found in Appendix \ref{app:G3}, as part of an in-depth case study on HOPV.

\subsection{Conclusion \& Future Work}
We represent molecules as random walks over an interpretable context-sensitive grammar over the motif graph, a hierarchical abstraction over the design space. We devise and execute a practical workflow that invites experts in the loop to enhance our design basis and representations by fragmenting molecules into well-established functional groups, creating a synergy between expert feedback and the quality of our representations. Our evaluation on downstream property prediction and molecular generation tasks shows our representation combines quantitative advantages in performance and efficiency with qualitative advantages of simplicity and enhanced interpretability. One promising avenue of future research is improving the autonomous extraction of interpretable grammar rules through learnable and/or human-guided approaches with Large Language Models.

%% file: body/6.appendix.tex
\section{Motif Collection Strategy} \label{app:A}
 Motifs are used to construct our motif graph $G$, which forms the design basis for both our generative and predictive methods. The complexity of our grammar as conveyed by the size of the motif graph $G$ for different motif collection strategies we tried are summarized in Table \ref{tab:complexity}. For the remainder of this section, we describe the Expert Annotation strategy which is our primary workflow. The strategies for obtaining motifs from literature and heuristic-based fragmentation are described in \ref{app:B1} and \ref{app:F1}, respectively.
\begin{table}[h!]
\centering
\begin{tabular}{l|l|l|l}
\toprule
Grammar Complexity $(|V|, |E|)$ & HOPV         & PTC          & GC         \\ \midrule
Literature                    & N/A          & N/A          & $(96, 3656)$ \\ \midrule
Expert                        & $(329, 37273)$ & $(407, 23145)$ & N/A        \\ \midrule
Heuristic                     & $(208, 16880)$ & $(279, 37968)$ & $(90, 4095)$ \\ \bottomrule
\end{tabular}
\caption{Number of nodes and edges of motif graph $G$ constructed using different annotation strategies}
\label{tab:complexity}
\end{table}
\subsection{Expert Annotation Workflow}
The workflow consists of two steps: molecule segmentation, and extracting the negative groups for pairwise attachments. Step 1 involves cooperation from an expert, and we detail our polished workflow to facilitate that process, which we have attempted with multiple experts. Step 2 can become automated after the expert identifies 1) governing rules for a particular dataset, and 2) important exceptions to the rule. On average, each dataset takes less than one working day for one expert to fully annotate and process. The annotated datasets for Group Contribution, HOPV, and PTC will be released upon publication.

\begin{figure*}[h!]
\centering
\includegraphics[width = \textwidth]{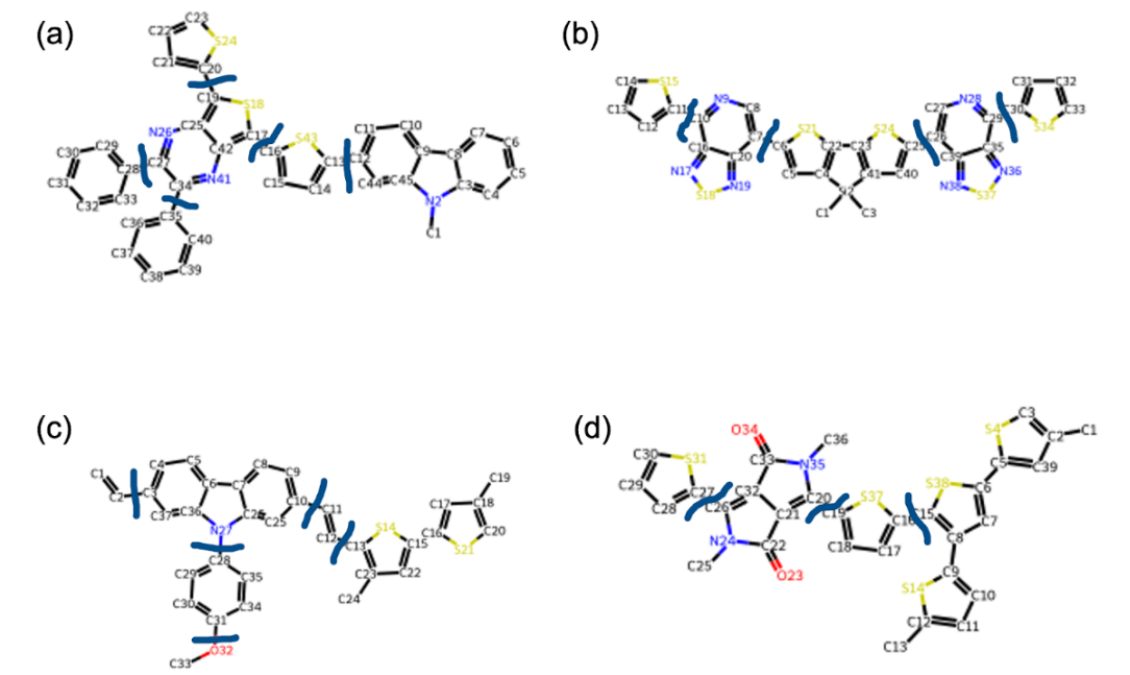}
\caption{Example segmentations for four molecules in the HOPV dataset. Segmentation locations are marked by the dark blue (teal) line.}
\label{expert-annotation}
\end{figure*}

\subsubsection*{Step 1: Expert Segmentation} \label{app:A1}

First, experts view figures of the molecules, and indicate the bonds to break in order to segment the molecule into coherently chosen sub-fragments, shown in Figure \ref{expert-annotation}. We provide a brief description of the example datasets we show here and elaborate on the rationale behind the experts' segmentation strategy:

\begin{table}[h!]
\centering
\caption{Segmentation of the molecules (a) to (d) in \ref{expert-annotation}. \textit{Bonds to break} indicates the chemical bonds to cut to create black fragments, while the \textit{black groups} and \textit{red groups} listed for each molecules correspond to one another, respectively.}
\begin{tabular}{c|c|c|c} \toprule
Structures & Bonds to Break & Black Groups & Red Groups\\\midrule
(a) & \shortstack{(12,13) (16,17) (19,20)\\ (27,28) (34,35)}  & \shortstack{(1,2,3,4,5,6,7,8,9,10,11,12,44,45) \\(13,14,15,16,43) (17,18,19,25,26,27,34,41,42) \\(20,21,22,23,24) (28,29,30,31,32,33) \\(35,36,37,38,39,40)} & \shortstack{(13) (12,17) (16,20,28,35)\\ (19) (27) (34)} \\\midrule
(b) & \shortstack{(10,11) (6,7) (25,26) \\ (29,30)} & \shortstack{(11,12,13,14,15) (7,8,9,10,16,17,18,19,20)\\(1,2,3,4,5,6,21,22,23,24,25,40,41) \\ (26,27,28,29,35,36,37,38,39)\\(30,31,32,33,34)} & \shortstack{(10) (11,6) \\ (7,26) (25,30) \\ (29)}\\\midrule
(c) & \shortstack{(2,3) (11,10) (12,13) \\ (27,28) (31,32)} & \shortstack{(1,2) (3,4,5,6,7,8,9,10,24,25,27,36,37)\\(28,29,30,31,34,35) (11,12) (32,33) \\ (13,14,15,16,17,18,19,20,21,22,23,24)} & \shortstack{(3) (2,11) \\ (27,32) (10,13)\\ (31) (12)} \\\midrule
(d) & (15,16) (19,20) (26,27) & \shortstack{(1,2,3,4,5,6,7,8,9,10,11,12,13,14,15,38,39)\\(16,17,18,19,37)\\(20,21,22,23,24,25,26,32,33,34,35,36)\\(27,28,29,30,31)} & \shortstack{(16) (15,20)\\(19,27) (26)}\\\bottomrule
    
\end{tabular}
\end{table}
\textbf{Predictive Toxicology Challenge (PTC)} \cite{Helma01}
The small molecules are characterized by distinct functional groups known for their carcinogenic properties or liver toxicity \cite{Miller49, Helma01}. These groups comprise a rich variety of elements such as halides, alkylating agents, epoxides, and furan rings. (\cref{examples}) Therefore, we specifically segmented it into functional groups and sub-structures that contribute most to the toxicity of the compounds \cite{Hughes15}.

\textbf{The Harvard organic photovoltaic dataset (HOPV)} \cite{hopv}
The process of segmenting the Harvard Organic Photovoltaic Dataset (HOPV15) demonstrates a methodical and efficient approach to categorizing photovoltaic data. This dataset contains a comprehensive collection of experimental photovoltaic data from literature coupled with quantum-chemical calculations across various conformers. The criteria for the extraction of the black group are clearly defined and systematically applied. Functional groups like vinyl, alcohol, ketone, aldehyde, amine, ester, and amide are separated as individual black fragments. Similarly, distinct black fragments are used for individual rings including benzene, pyrrole, and thiophene, in acknowledgement of their Pi-orbital electron delocalization. Complex structures with multiple consecutive rings, known for their distinctive HOMO-LUMO bandgaps and electrochemical properties, such as thieno[3,4-b]pyrazine, carbazole, and 2,5-dimethyl-3,6-dioxo-2,3,5,6-tetrahydropyrrolo[3,4-c]pyrrole, are also segmented as individual entities. Moreover, for groups of 2-3 consecutive symmetrical thiophene or pyrrole units, the methodology captures the significance of maintaining them as a complete black group because these consecutive groups sustain the electron cloud delocalization between repeating units, strongly influencing optical and electronic properties not limited to light absorption, charge transport, and luminescent properties in photovoltaic applications. Meanwhile, this method of segmentation enhances utility and understanding of the results by clearly basing predictions on existing important structures.

\paragraph{Defining Membership.} The Membership metric is reported in \ref{sec:generation} after further consultation with experts, who identify the presence of Thiophene as a proxy for Membership to HOPV, and the presence of Chloride/Bromide Halides (a key indicator of toxicity) for PTC. In the case of both datasets, the Membership metric is only a sanity check that the method can produce a non-trivial number of characteristic compounds. Here's our justification for the criteria on each dataset:
\begin{enumerate}[leftmargin=*,topsep=0pt,noitemsep]
    \item A chloroalkane (Cl-C) is the most common motif in the PTC dataset. Yet, it is still not present in a majority of structures, making the broader class of alkyl halides (Cl-C, Br-C-C) the best choice for a membership criterion. Their prevalence is attributed to their reactivity and ability to undergo metabolic activation \cite{leung2012}, leading to the formation of highly reactive intermediates that can interact with DNA and other cellular components, potentially initiating carcinogenic processes. Although not all carcinogenic compounds will necessarily contain this class of motifs, their presence contributes a strong likelihood.
    \item Thiophene, a 5-member ring with one sulfur group, is the most common motif in the HOPV dataset, making it the best choice for a single-motif membership criterion for HOPV. More broadly, thiophene and its derivatives are arguably the most common chemical substructure in photovoltaics due to their ability to donate electrons, resulting in particularly high highest occupied molecular orbital (HOMO) levels, along with stability, tunability of energy levels, and compatibility with film forming techniques. While not every suitable organic photovoltaic compound will contain it, the vast majority will.
\end{enumerate}

In both cases, our method can easily achieve 100\% membership with a slight modification to the sampling procedure: instead of iterating through every possible starting motif node, we always initialize our random walk at the membership motif. We choose not to modify our sampling procedure, and instead include this metric in Table \ref{molgen} for completeness, since it is still a good sanity check for other methods to show they generate a non-trivial fraction of candidates with those motif(s).

\subsubsection*{Step 2: Extracting Red Groups}\label{app:A2}
Key to the definition of our motif graph is the specification of red groups $(v_R \subset v)$ that define the possible pairwise attachments between motifs. There are no hard rules, but generally red groups should be minimally necessary. It should be: 1) consistent, for enabling more attachments, hence making the motif graph denser; 2) small, for enabling fast isomorphism checking during the precomputation of the motif graph, and 3) necessary, ensuring only valid attachments. Failure to follow 3) can generate chemically disallowed structures.


\begin{table}[h]
\centering
\caption{Context determination rules and examples on datasets Group Contribution, HOPV and PTC.}
\begin{tabular}{c|c|c}\toprule
    Dataset & Rule & Example  \\\midrule
    Group Contribution &  \shortstack{We directly use the released groups \\ in \cite{group-contrib-1, group-contrib-2}.} &

\includegraphics[width = \textwidth / 3]{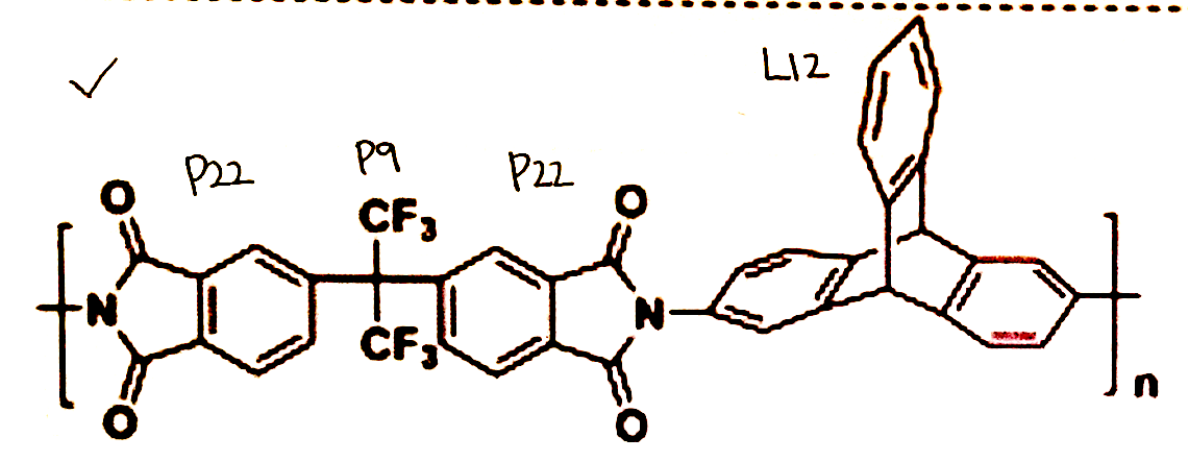}

\\\midrule
    HOPV & \shortstack{- For groups of a single atom -- pick\\ ring of neighbor fragment if possible\\
- For groups of multiple atoms -- pick\\ only the connected atoms in the neighbor fragment} & \includegraphics[width = \textwidth / 4]{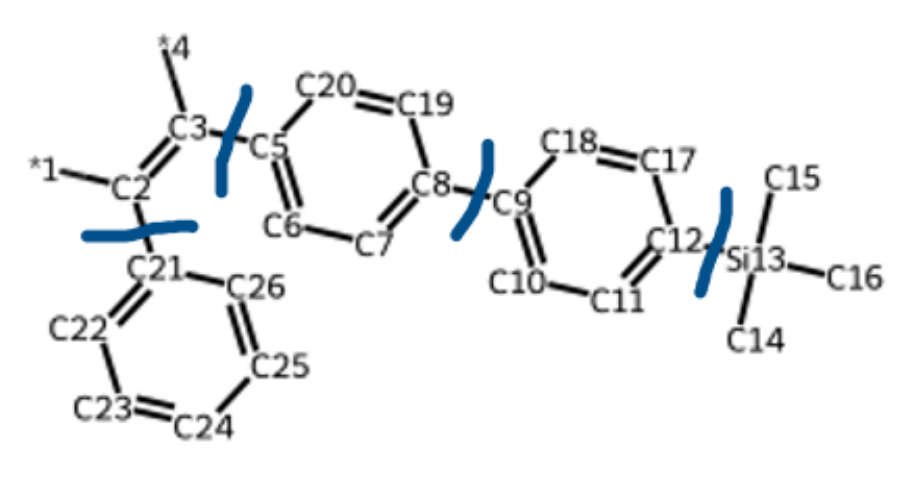}  \\\midrule
    PTC & - Same as HOPV & \includegraphics[width = \textwidth / 3]{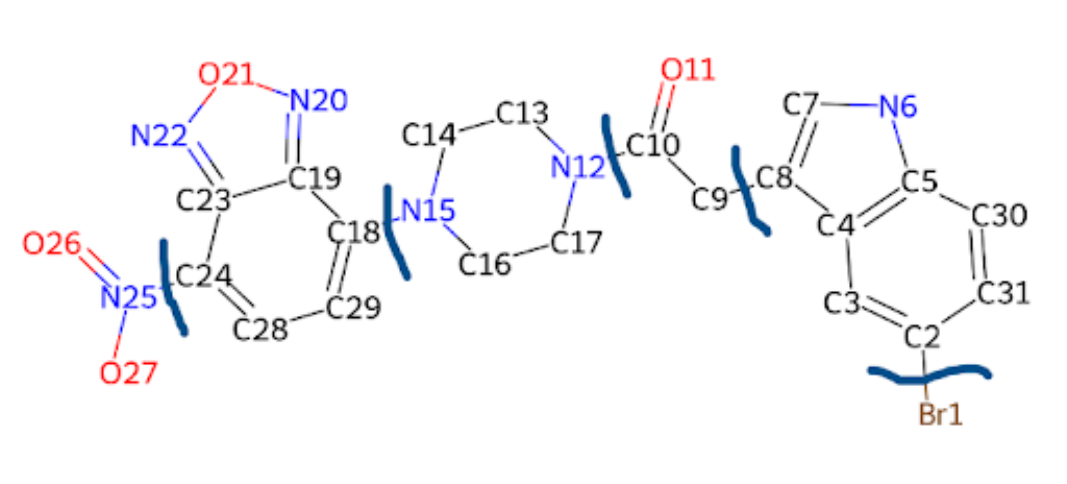}    \\\bottomrule
\end{tabular}
\end{table}

\section{Representing Existing Molecules as Walks on This Graph} \label{app:B}

\subsection{Extracting Walks from Segmentation (HOPV, PTC)} \label{app:B1}
During segmentation, we use the workflow in Appendix \ref{app:A2} to segment a molecule into fragments. In doing so, we also obtain the molecule’s representation $H_M$ as a directed subgraph over the motif graph. The pseudocode is found in \cref{algo_extract_walk}.

\begin{algorithm2e}[h!]
\LinesNumbered
\newcommand\mycommfont[1]{\footnotesize\ttfamily\textcolor{blue}{#1}}
\SetCommentSty{mycommfont}
\caption{function extract\_walk(D,B)}\label{algo_extract_walk}
\setcounter{AlgoLine}{0}
\SetKwInput{Initialization}{Input}
\Initialization{$D = [M_i \mid i = 1, \ldots, |D|]$ \tcp{dataset of molecules}}
$B = [B_i \mid i = 1, \ldots, |D|]$; \tcp{annotation, i.e. bonds to break, for each molecule}
$D_G = []; V = \{\}; H= [];$\\
\For{{\upshape $i \text{ in range }(\text{len}(D))$}}
{
$F_i \gets \text{break\_bonds}(M_i, B_i)$; \tcp{break bonds and form fragments}
$G_i \gets \text{form\_graph}(F_i,B_i)$; \tcp{graph of motifs, edges preserving connections}
\For{{\upshape $f_1 \text{ in } F_i$}}
{
\For{{\upshape $f_2 \text{ in } N_{F_i}(f_1)$}}
{
b $\gets G_i$.edges[(f1,f2)]; \tcp{bond(s) connecting $f_1, f_2$}
$\text{rule} \gets \text{apply\_rule}(f_1,f_2,b)$; 
\\
V.add(rule);
}
}
$D_G.\text{append}(G_i)$; 
}
\For{{\upshape $G_i \text{ in }  D_G$}}
{
$H_i \gets \text{traverse\_dag}(G_i, G)$;\\
H.append($H_i$);
}
Out: H,G
\end{algorithm2e}

\begin{algorithm2e}[h!]
\LinesNumbered
\newcommand\mycommfont[1]{\footnotesize\ttfamily\textcolor{blue}{#1}}
\SetCommentSty{mycommfont}
\caption{function traverse\_dag($G_i, G$)}\label{algo_traverse_dag}
\setcounter{AlgoLine}{0}
\SetKwInput{Initialization}{Input}
\Initialization{$G_i$, $G$, $N_{G_i}$} \tcp{graph of fragments, motif graph, neighbor iterator}
paths $\gets$ all\_pairs\_shortest\_paths($G_i$);\\
path\_len = 0;\\
\For{{\upshape src in paths}}
{
\For{{\upshape dest in paths[src]}}
{
    \If{ \upshape len(paths[src][dest]) $>$ path\_len}
    {
        path\_len $\gets$ paths[src][dest];\\
        longest\_path $\gets$ paths[src][dest];
    }
}
}
visited $\gets$ \{\};\\
visited[src] $\gets$ True;\\
root $\gets$ Node(src, main = (src in longest\_path));\\
Q $\gets$ queue([(root,src)]);\\
\While{\upshape !Q.empty()}
{
prev\_node, prev $\gets$ Q.dequeue();\\
\For{{\upshape cur in $N_{G_i}$(prev)}}
{
\If{\upshape visited[cur]}
{
continue
}
$e \gets G_i$.edges[(prev, cur)];\\
e\_index $\gets$ find\_edge(e, G.edges[(prev,cur)]);\\
cur\_node $\gets$ Node(cur, main = (cur in longest\_path));\\
prev\_node.add\_child(cur\_node, e\_index);\\
vis[cur] $\gets$ True;\\
Q.enqueue((cur\_node, cur))
}
}
Out: root
\end{algorithm2e}

\cref{algo_traverse_dag} linearizes the molecule into a directed acyclic graph (DAG). This procedure begins by finding the longest path, and choosing a consistent ordering over neighbors ($N_{G_i}$) to determine the random walk sequence. We elaborate on the reasoning behind this canonicalization in Appendix \ref{app:B4}. The DAG constraint enables our graph diffusion process to become a generator of new molecules (as will be discussed in Appendix \ref{app:D}), in addition to capturing the distribution of existing ones. Thus, we specifically ask experts to create segmentations that are acyclic, which they naturally do in nearly all cases anyway. In the case of monomers, this canonicalization is consistent with the IUPAC nomenclature\cite{iupac} of linearizing a monomer via its longest (main) chain, where $N_{G_i}$ should iterate over neighbor fragments that descend side chains before the consecutive fragment on the backbone of the main chain. More specifically, \text{src} and \text{dest} in \cref{algo_traverse_dag} correspond to the first and last group of the main chain.

\subsection{Extracting Walks From Literature (Case Study of Group Contribution)} \label{app:B2}



The Group Contribution dataset includes a compilation of motifs characterized for gas separation, including common organic chemical functional groups as well as important scaffold functional groups such as Triptycene and its derivatives, dioxin and its derivatives, and N-methylphthalimide and PIM-1 and its derivatives \cite{microporous, Wang20}. These functional groups contribute significantly to maintaining the structures and properties of 3D scaffold building blocks in polymer self-assembly, which in turn play a significant role in gas separation processes, i.e. the separation of $H_2, H_2/N_2, O_2, O_2/N_2, CO_2, CO_2/CH_4$ which are common separation tasks important in gas and oil industry. The steps we take for compiling this dataset of segmented monomers are as follows:
\begin{enumerate}[leftmargin=*,topsep=0pt,noitemsep]
    \item We obtain an established compilation of groups \cite{group-contrib-1, group-contrib-2} for microporous polymers. 
    \item We visually segment the monomers in \cite{microporous} into random walks over the groups identified in Step 1. 
    \item We collect experimental permeability and separation performances for 114 of the monomers identified in Step 2.
\end{enumerate}

In addition to the motifs used here, the concept of such segmentation arises naturally across other application domains in chemical design. Within synthetic organic chemistry, molecular design plays a governing role in advancing new technology \cite{ChemicalDesignSemiconductors}. Understanding of the behavior of a molecule or polymer in an application is commonly described by experts using the function of key subparts, particularly key functional groups, scaffold structures, and backbone architectures within a molecule or monomer, and their arrangement relative to each other, rather than considering atom-by-atom or a molecule as a whole. In chemical design, new molecules can be complex and, when designed by hand in traditional ways, are built from these relatively modular subcomponents. This approach naturally takes advantage of the physical laws by which molecules are built by synthesis reactions, where a discrete set of additions and substitutions are allowed to finally construct a desired target structure. Such methods of chemical design find broader application in drug discovery for pharmaceuticals, surfactant and detergent design \cite{surfactants, surfactants2}, organic semiconductors \cite{ChemicalDesignSemiconductors}, photoinitiators \cite{photoinitiators}, and more recyclable plastics \cite{epoxy}, among other uses, in each of which chemists fine-tune properties of such components by adjusting the selection and arrangement of these sub-structures, or otherwise use them as a guide for understanding performance. 

Utilizing groups from existing structures as well as discovery of new and novel structures, researchers can predict performance, find new uses for existing molecules, discover new molecules, and further optimize structures for better performance. Utilizing machine learning models has been show to drastically decrease the time and cost of such methods while simultaneously improving throughput by creating and screening novel structures in a single step and providing researchers with predictions of target molecules that have higher potential for success, which are then verified by experts. As presented by \cite{group-contrib-2}, different structural elements and functional groups present in effective drug molecules can be identified and recombined in new architectures. These novel structures can then be tested using computer models to benchmark likely efficacy given new targets or modifications to binding sites.

\subsection{Graph Augmentation} \label{app:B3}

The motif graph is the directed, multi-edge graph $G=(V,E)$. When traversing to a previously seen motif v, there is ambiguity in whether the random walk forms a cycle vs creating a copy of the previous motif and appending to the trajectory. To remove this ambiguity, the random walk traverses a duplicate node, $v_k$ for the latter case. A dataset of molecules and their representations, $D:=\{(M, H_M)\}$ thus induces ``an augmented version of $G$" =: $G’$. For each $v \in V$, let $K_v=\max(\text{count}(v, H_M)$ for M). We create duplicates for $v$ and the in/out-edges of $v$: 
\begin{align}
    V’ &\leftarrow V \cup \bigcup_{v\in V} \{v_{k} \mid k=0,\ldots,K_v-2\} \\
    E’ &\leftarrow E \cup \bigcup_{v\in V} \{(v_k, v’, e) \mid (v, v’, e) \in E, \forall k=0,\ldots,K_v-2\} \\
    E’ &\leftarrow E \cup \bigcup_{v\in V} \{(v’, v_k, e) \mid (v’, v, e) \in E, \forall k=0,\ldots,K_v-2\}.
\end{align}

Molecule $M=(V_M,E_M)$ is then a rooted subgraph of $G’$. In the main text, we refer to $G$ as its augmented version, to simplify the notation.

\subsection{Data Augmentation} \label{app:B4}

Like the Simplified molecular-input line-entry system (SMILES), our description, $\hat{H}_M$, of a molecule is not unique. We tried, to varying extents, balancing between canonicalizing the description vs applying data augmentation during the grammar training phase. 

As described in Algorithm \ref{algo_traverse_dag}, we linearize a molecule by first setting its “main chain” -- the longest shortest path of $H_M$. If this happens to be part of a cycle, we disregard one edge. If there are multiple longest shortest paths, we choose the one whose first differing node comes first in our canonical ordering over the nodes of G.

We tried two types of data augmentation:
\begin{enumerate}[leftmargin=*,topsep=0pt,noitemsep]
    \item Reversing the direction of the main chain. 
    \item For each node, trying every permutation over the side chains descending from it.
\end{enumerate}
However, we noticed no practical improvements in training loss or downstream task performance when either of the two types of augmentation were applied. We believe that, given our parameter estimation procedure, the consistently applied canonicalization over the nodes of $G$ improves data-efficiency by significantly reducing the hypothesis space.

\section{Building the Motif Graph \& More Related Works} \label{app:C}

Expanding on Section \ref{sec-method-1}, we apply a subgraph-matching algorithm with pseudocode in Algorithm \ref{build_motif_graph} over all pairs of motifs $v_1, v_2$. This algorithm is embarrassingly parallel and runtime-efficient as the subgraph $v_{R_l}$ is, unless specified otherwise, a few atoms or a ring. RDKit\cite{rdkit} provides out-of-the-box implementations for subgraph matching optimized for molecular sub-fragments like rings, enabling a significant speedup in runtime.

\begin{algorithm2e}[h!]
\LinesNumbered
\newcommand\mycommfont[1]{\footnotesize\ttfamily\textcolor{blue}{#1}}
\SetCommentSty{mycommfont}
\caption{function build\_motif\_graph(V)}\label{build_motif_graph}
\setcounter{AlgoLine}{0}
\SetKwInput{Initialization}{Input}
\Initialization{V} \tcp{motifs}
G = graph(V); \\
\For{{\upshape $v_1$ in $V$}}
{
\For{{\upshape $v_2$ in $V$}}
{
\For{{\upshape $l1$ in $v_{1_R}$}}
{
sub\_${2_r} \gets$ extract\_subgraph$(v_2, v_{1_{R_l1}})$;\\
b2\_sub $\gets$ substruct\_matches$(v_2, \text{sub\_}{2_r}$;\\
b2\_all $\gets$ isomorphisms\_iter$(v_2,$ b2\_sub);\\
conn\_b1 = [];
\For{{\upshape b1 in b1\_all}}{
\If{\upshape connected$(v_1(v_{1_{R_l1}} + b1))$}{
conn\_b1.append(b1);
}
}
\For{{\upshape l2 in $v_{2_R}$}}
{
sub\_$1_r \gets$ extract\_subgraph$(v_1, v_{2_{R_l2}})$;\\
b1\_sub $\gets$ substruct\_matches$(v_1,$ sub\_${1_r}$;\\
b1\_all $\gets$ isomorphisms\_iter$(v_1, b1$\_sub);\\
conn\_b2 = [];\\
\For{{\upshape b2 in b2\_all}}
{
\If{\upshape connected($v_2(b2 + v_{2_{R_l2}}))$}{
conn\_b2.append(b2);
}
}
\For{{\upshape b2 in conn\_v2}}{
\For{{\upshape b1 in conn\_b1}}{
sub\_1 $\gets v_1(v_{1_{R_l1}} + b1)$;\\
sub\_2 $\gets v_2(v_{2_{R_l2}} + b2)$;
\If{\upshape isomorphic(sub\_1, sub\_2)}{
$e_{l1,l2} \gets (v_1, v_2,$ {r\_grp\_1: $v_{1_{R_l1}},$ r\_grp\_2: $v_{2_{R_l2}}$, $b_1$: $b_2$: $b_2$});\\
G.add\_edge(e\_$\{l1,l2\}$);
}
}
}
}
}
}
}
Out: G
\end{algorithm2e}

\begin{figure*}[h!]
\centering
\includegraphics[width = \textwidth]{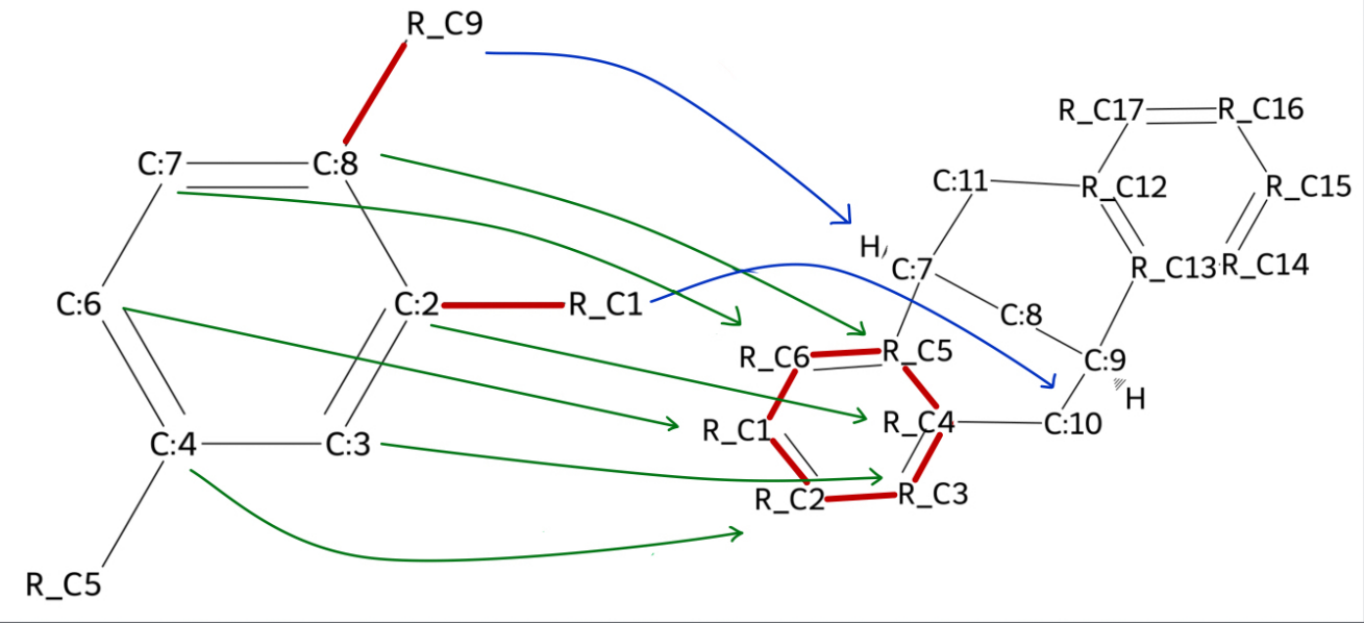}
\caption{Following notations of main text, $\{v_1\}_{r_1} = \{1,9\},  \{v_1\}_{r_2} = \{1\}, \{v_1\}_{r_3} = \{9\}, \{v_1\}_{r_4} = \{5\}$. $\{v_2\}_{r_1} = \{1,2,3,4,5,6\}, \{v_2\}_{r_2} = \{12,13,14,15,16,17\}$. We annotate $e_{1, 1}$, where $b_1=\{6,4,3,2,8,7\}, b_2=\{10,7\}$ provides the “certificate” of a successful match.}
\label{match-example}
\end{figure*}

\subsection{Connection to Dual Graph of Geo-DEG's Meta-Grammar} \label{app:C1}

Our proposed directed multi-digraph also conceptualizes the dual graph of the Geo-DEG meta-geometry. The essence of the Geo-DEG meta-geometry lies in its completeness, a characteristic inherently inherited by our proposed digraph. A significant advantage of our approach is the substantial reduction in complexity. To elucidate this process, consider the construction of our multi-digraph from the Geo-DEG meta-geometry, denoted as $G_g=(V_g, E_g)$. The initial step involves replacing each node in $V_g$, which represents a junction tree, with all feasible molecule structures derived from motifs that maintain the same junction tree structure. Subsequently, we augment $E_g$ with fully connected edges between these sets of molecule structures. The dual graph, $G_dg$, is then derived from $G_g$, where each node from $V_g$ is transformed into an edge, and each edge from $E_g$ becomes a node. This dual graph not only preserves the completeness of the original graph but also provides an intuitive representation of molecular assembly. Each node in the dual graph symbolizes a motif, and traversing this graph illustrates the process of assembling a molecule by adding motifs. To refine this representation, we eliminate duplicate nodes in the dual graph, ensuring each node's uniqueness.

The representation's completeness is maintained because every possible molecule structure derivable from the motifs is accounted for in the dual graph. Each pathway through the graph represents a unique assembly sequence of motifs, translating into a distinct molecular structure. The reduction in complexity arises from the transformation process. By converting the original graph into its dual form, we reduce the granularity of representation. Instead of representing every possible molecular structure as a separate node, we represent them as pathways through the dual graph. This approach significantly decreases the number of nodes and edges required, leading to a more manageable yet complete representation of the molecular structures.

\subsection{Connection to Graph Coarsening} \label{app:C2}
Mathematically, the motif graph advocated in this work is the \emph{quotient graph} of the molecular graph, under the equivalence relation defined as $u \equiv v$ if nodes $u$ and $v$ belong to the same motif. As our motifs do not overlap and jointly cover all nodes of the molecular graph, they define a partitioning of the graph. In scientific computing, collapsing each partition into a single node and retaining edges crossing partitions is called \emph{graph coarsening}, which is a commonly used technique to solve large-scale problems, notably solving sparse linear systems of equations \cite{Chen2022}. Working on the coarsened version of the graph (i.e., the quotient graph) is computationally attractive as the graph size is much smaller. Moreover, when applied to machine learning problems such as graph classification, it is demonstrated that the representation learned from the quotient graph can be as predictive as that learned from the original graph \cite{Chen2023, Ma2021, Cai2021}. Favorably, a unique scenario of this work is that all concerned (molecular) graphs share the same set of motifs, which brings in the potential benefit of learning better molecule representations based on motif representations that form the basis of all molecules.

\subsection{Connection to Random Walk Literature} \label{app:C3}
Our parameterization of the random walk is by learning a graph heat diffusion process over the motif graph $G$. The relationship between graph heat diffusion and random walk has been studied before \cite{masuda2017}, but we integrate two new ideas: 1) making the Laplacian (edge weights) learnable and dynamically adjustable, and 2) conditioning the adjustment on an order-invariant memory. The justification as to why we don't just use autoregressive models is part of a larger discussion on the respective merits of autoregressive models vs grammar-based approaches. In data-efficient settings, previous works \cite{deg, ours} show grammar (esp. context-free grammar) work well due to the relatively small (tens/hundreds) number of examples needed to learn valid rules and derivation sequences. Meanwhile, the number of possible hidden states that autoregressive models \cite{li2018, you2018, liu2018molecule} are parameterized to learn is exponential (to the length of the sequence), and learning a good parameterization is difficult \cite{jtvae, hgraph2graph}. We take a middle ground, combining the data-efficient advantages of context-free grammar and the expressivity of autoregressive models, by introducing a context-sensitive grammar which utilizes a set-based memory during the random walk. The set-based memory mechanism $c^{(t)}$ keeps an order-invariant memory of the nodes visited so far. Without the memory mechanism, our model becomes an order 1 Markov process. Previous literature show that higher-order random walks are required to capture temporal correlations in edge activations \cite{rosvall2014, masuda2017}, with a tradeoff of complexity and practicality. In the design of complex and modular structures, the order 1 Markov assumption is not sufficient (see footnote \ref{footnote:2} in the paper). Meanwhile, higher-order models make it difficult to scale our grammar to larger motif graphs. We take a middle ground by introducing a set-based memory state, replacing the entire visit history with a summary of node visit counts. In particular, prior works study how memory mechanisms in random walks affect exploration efficiency \cite{fang2023, gasieniec2008} and enable negative/positive feedback \cite{fang2023, pemantle1988}. Our results in Section \ref{sec:generation} demonstrate the efficacy of this approach.

\section{Grammar Learning} \label{app:D}

\subsection{Graph Diffusion Strategy} \label{app:D1}

Our strategy is to encode the dataset of walk trajectories by training the parameters of our graph diffusion process to recover the ground-truth state of a particle being diffused over the motif graph. We use stochastic gradient descent and choose between a “forcing” approach (where a single particle transitions from one state to another) and a “split” approach (where a single particle splits its mass equally along the out-edges of its current state). See the pseudocode in \cref{algo_diffusion}.

\begin{algorithm2e}[h!]
\LinesNumbered
\newcommand\mycommfont[1]{\footnotesize\ttfamily\textcolor{blue}{#1}}
\SetCommentSty{mycommfont}
\caption{function re\_order(childs)}\label{re_order}
\setcounter{AlgoLine}{0}
\SetKwInput{Initialization}{Input}
\Initialization{childs \tcp{children}}
ordered\_childs $\gets$ sorted(childs, key = $\lambda$ c: (c.main, c.id));\\
Out: ordered\_childs \tcp{re-ordered children, with side-chain descendants first}
\end{algorithm2e}

\begin{algorithm2e}[h!]
\LinesNumbered
\newcommand\mycommfont[1]{\footnotesize\ttfamily\textcolor{blue}{#1}}
\SetCommentSty{mycommfont}
\caption{function dfs\_walk(cur, traj)}\label{dfs_walk}
\setcounter{AlgoLine}{0}
\SetKwInput{Initialization}{Input}
\Initialization{cur, traj \tcp{children}}
traj.append(cur);\\
childs $\gets$ re\_order(cur.children);\\
\For{{\upshape c in childs}}
{
cur\_len $\gets$ len(traj);\\
dfs\_walk(c, traj);\\
\If{\upshape !c.main}
{
traj $\gets$ traj + reverse(traj[cur\_len:]);\\
}
}

\end{algorithm2e}

\begin{algorithm2e}[h]
\LinesNumbered
\newcommand\mycommfont[1]{\footnotesize\ttfamily\textcolor{blue}{#1}}
\SetCommentSty{mycommfont}
\caption{function algo-diffusion}\label{algo_diffusion}
\setcounter{AlgoLine}{0}
\SetKwInput{Initialization}{Input}
\Initialization{T, G, alpha, strategy \tcp{number of time-steps, motif graph, learning rate, either 'split' or 'forcing'}}
E $\gets$ rand($|G| \times |G|$);\\
W $\gets$ rand($|G|, |G|*|G|$);\\
b $\gets$ zeros($|G|$);\\
\For{{\upshape $(H_M, E_M)$ in D}}
{
$c^{(0)} \gets$ [0 for v in G];\\
$x^{(0)} \gets$ [1 if v==$H_M$.root else 0 for v in G];\\
$p^{(0)} \gets$ [1 if v==$H_M$.root else 0 for v in G];\\
\If{\upshape strategy == 'forcing'}{
traj $\gets$ [];\\
dfs\_walk($H_M$.root, traj);
}
\For{{\upshape $t=0, \ldots, T-1$}}
{
$c^{(t+1)} = \frac{t}{t+1} \cdot c^t + \frac{1}{t+1} \cdot p^t$;\\
$W^{(t+1)} = E + f(c^{(t+1)})$;\\
$x^{(t+1)} = x^t + (D-W^{(t+1)})x^t$;\\
\If{\upshape strategy == 'forcing'}{
$p^{(t+1)} \gets$ [1 if v == traj[(t+1)\%len(traj)] else 0 for v in G];
}
\Else{\For{{\upshape i in G}}{
$p^{(t+1)}_i \gets \sum_{(j,i) \in E_M} \frac{p^t_j}{d_j}$;
}}
Loss $\gets$ MSE($x_t$, $p^t$);\\
E $\gets$ E - $\frac{\text{dLoss}}{\text{dE}}$;\\
W $\gets$ W - $\alpha * \frac{\text{dLoss}}{\text{dE}}$;\\
b $\gets$ b - $\alpha *  \frac{\text{dLoss}}{\text{db}}$;
}
}
Out = E,W,b
\end{algorithm2e}

\subsection{Visualizing Learning Process} \label{app:D2}

\begin{figure*}[h!]
\centering
\includegraphics[width = \textwidth]{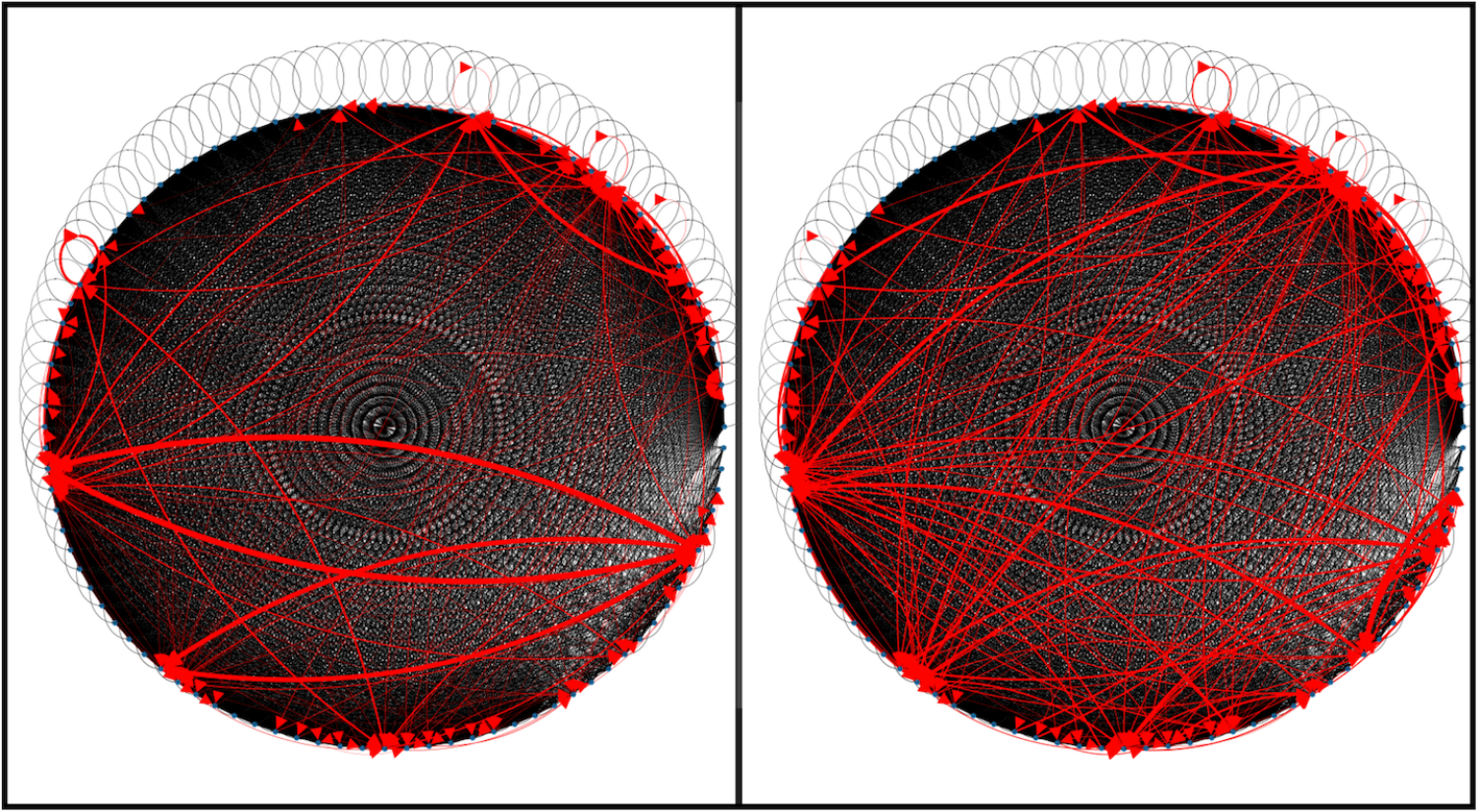}
\caption{(Left) The raw data of Group Contribution. The edge thickness is proportional to the number of monomers whose random walk representations traverse the edge.
(Right) The learned parameter matrix of E after training converges The grammar both retains essential nodes and edges and smoothens the distribution of edge weights.}
\label{E-evolution-group-contribution}
\end{figure*}

\begin{figure*}[h!]
\centering
\includegraphics[width = \textwidth]{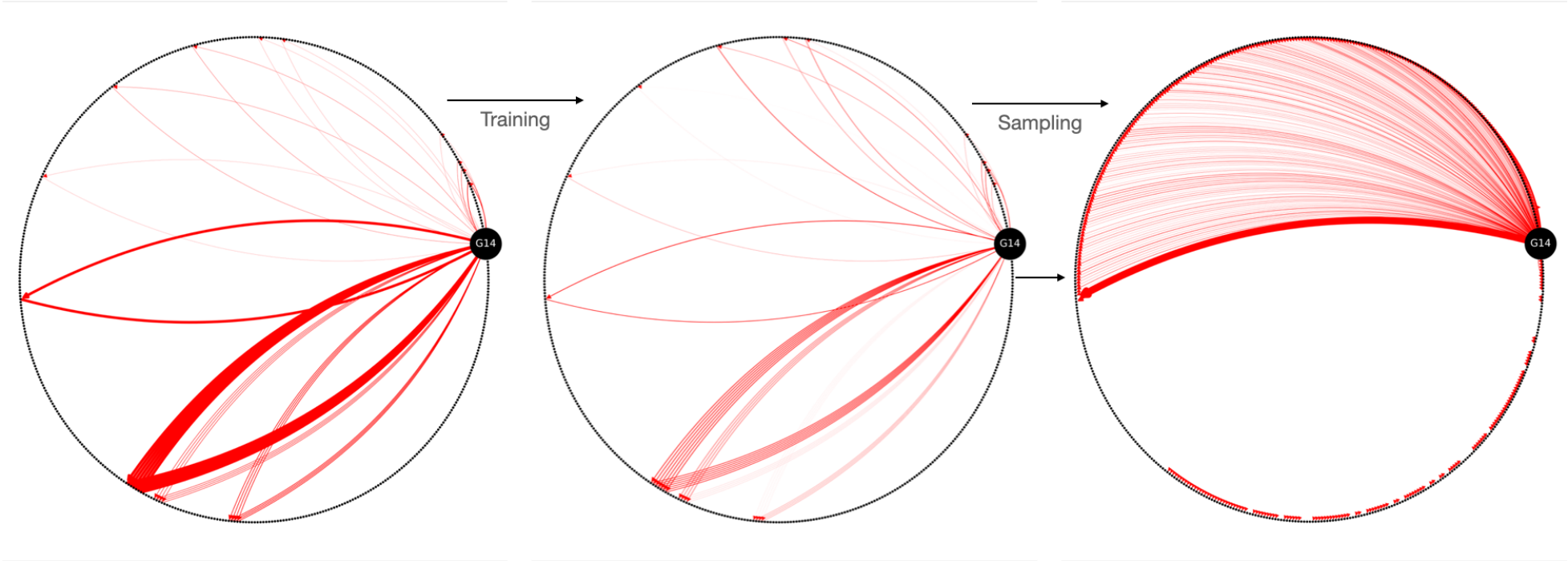}
\caption{We show the weight evolution of the edges incidental to $G14$ on HOPV. (Left) After processing the raw dataset into random walks, we visualize the empirical distribution of edge traversals. (Middle) After learning our context-sensitive grammar, we plot the prior edge weights, i.e. the memory-free parameter $E$. (Right) We plot the transition probabilities starting at $G14$ during the random walk generation process.}
\label{E-evolution-example}
\end{figure*}

In Figure \ref{E-evolution-group-contribution} and Figure \ref{E-evolution-example}, we see our grammar’s capacity to estimate the prior edge weights, $E$, through training, as well as correct the edge weights via a memory-sensitive adjustment during the random walk. Weights of edges that are commonly traversed after $G14$ will be amplified during training, and weights of edges that visit $G14$ from another state will be diminished.

\section{Property Prediction} \label{app:E}
\subsection{Graph Neural Network Design Choices} \label{app:E1}

\begin{table}[h!]
\centering
\caption{Hyperparameter settings for property prediction}
\label{tab:gnn-design-choices}
\begin{tabular}{l|l}\toprule
\textbf{Hyperparameter} & \textbf{Value}\\ \midrule
Number of layers & 5\\\midrule
Activation & ReLU\\\midrule
Hidden dimension & 16\\\midrule
Motif featurization & Morgan fingerprint\\\midrule
Motif feature dimension & 2048 \\\midrule
Input feature dimension \footnotemark & 2048 + $2048$ + $|G|$\\\midrule
Batch Size & 1\\\midrule
Learning Rate & 1e-3\\\bottomrule
\end{tabular}
\end{table}
We apply a Graph Isomorphism Network \cite{gin} with hyperparameters in Table \ref{tab:gnn-design-choices}. For molecule M with representation $H_M$, the node-level features include: a) the Morgan fingerprint of the motif $v_i$ (dimension 2048), b) the memory-free weights of its out-edges (dimension $|G|$), i.e. E[i]. We also concatenate the Morgan fingerprint of M. 

\subsection{Bag-of-Motifs Design Choices} \label{app:E2}

We obtain $|G|$-dimension motif-occurrence feature vector for each $M$. Similar to Ours, we concatenate the Morgan fingerprint of $M$ to it. We use XGBoost with $16$ estimators (boosting rounds) and maximum tree depth of $10$.



\subsection{Optimization Design Choices} \label{app:E3}

We apply the Adam optimizer with stochastic gradient descent. To mitigate noisy training dynamics, we report the mean and standard deviation over $3$ runs, corresponding to $3$ random seeds during data splitting. We initialize weights using the Gaussian distribution.

\section{Generating Novel Random Walks} \label{app:F}
\begin{figure}[h!]
  \centering
  \includegraphics[width=0.35\textwidth]{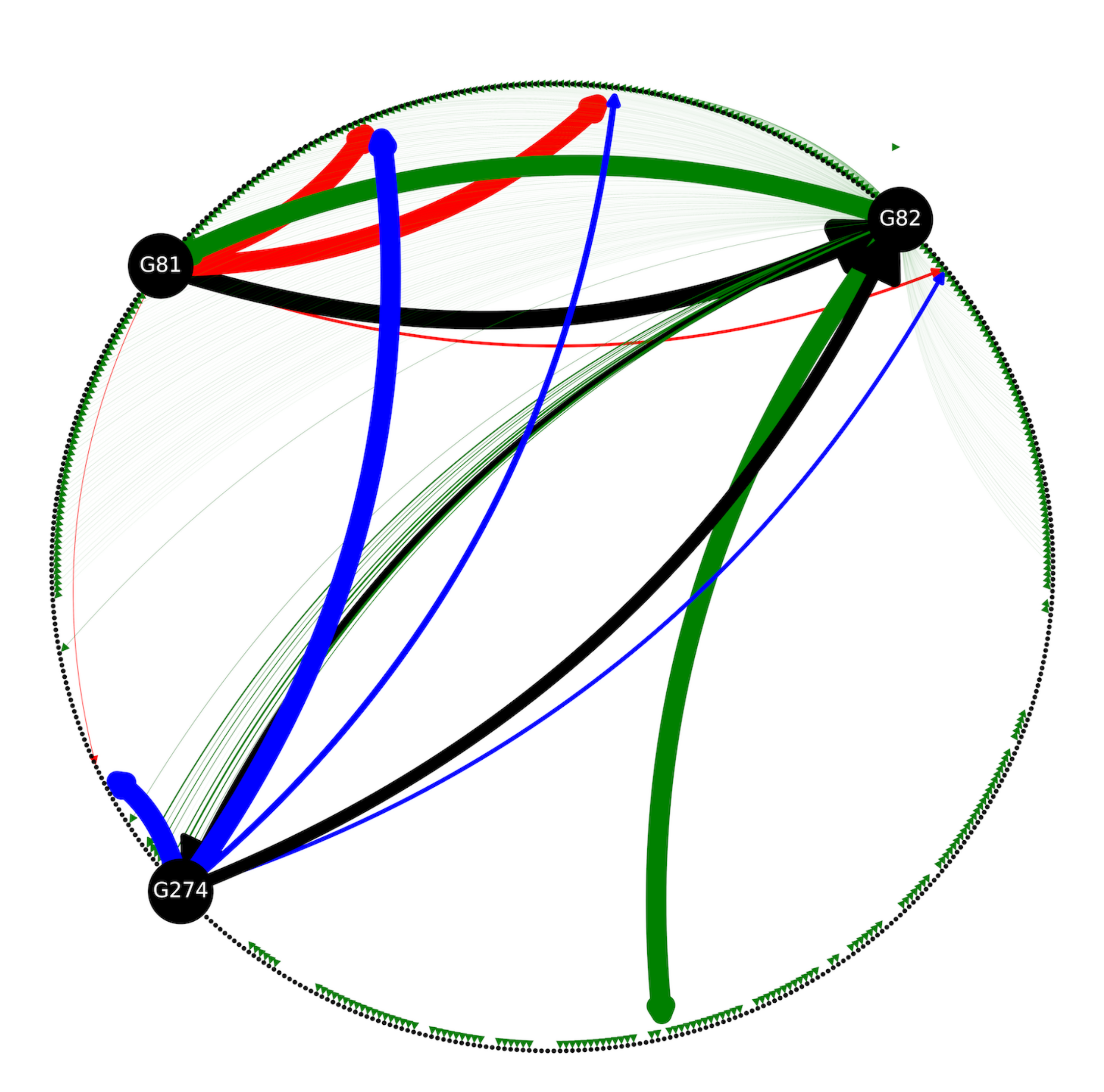}
  \caption{Generation of a random walk $G81 \rightarrow G82 \rightarrow G274 \rightarrow G82:1$; The possible transitions from $G81$, $G82$ and $G274$ are in \textcolor{red}{Red}, \textcolor{teal}{Green}, and \textcolor{blue}{Blue} (with thickness proportional to probability).}
  \label{fig:walk-3}
\end{figure}

We illustrate the generation of the random walk with notation $G81 \rightarrow G82 \rightarrow G274 \rightarrow G82:1$ in Figure \ref{fig:walk-3}. Our graph resolves any ambiguity of whether to revisit $G82$ or attach a new copy of $G82$ to the molecule by attaching a colon for each newly attached motif that has a naming conflict. This is possible after augmenting $G$ with duplicates of the motif (see Appendix \ref{app:B3}), which, in practice, has a negligible increase the complexity of $G$.

\begin{algorithm2e}[h!]
\LinesNumbered
\newcommand\mycommfont[1]{\footnotesize\ttfamily\textcolor{blue}{#1}}
\SetCommentSty{mycommfont}
\caption{function generate}\label{algo2}
\setcounter{AlgoLine}{0}
\SetKwInput{Initialization}{Input}
\Initialization{G \tcp{motif graph}}
root\_M $\sim$ V; \tcp{can sample according to prior}
loop\_back; \tcp{whether to loop back (applies for monomers)}
root $\gets$ Node(root\_M);\\
H $\gets$ root;\\
M $\gets$ molecule(root\_M); \tcp{initialize the molecule}
t $\gets 0$;\\
$c^{(t)} \gets$ [0 for v in V];\\
terminate $\gets$ False;\\
\While{\upshape !terminate}
{
$p^{(t)} \gets$ [1 for v in V if v==H.val else 0];\\
$c^{(t+1)} \gets \frac{t}{t+1} \cdot c^{(t)} + \frac{1}{t+1} \cdot p^{(t)}$;\\
$W^{(t+1)} \gets$ E + f($c^{(t+1)}$);\\
$x^{(t+1)} \gets x^{(t)} + $ (D-$W^{(t+1)}x^{(t)}$);\\
mask\_attach, mask\_return $\gets$ mask\_possible(M,G,H);\\
mask $\gets$ mask\_attach $|$ mask\_return;\\
$x^{(t+1)} \gets $ $\frac{\text{mask}*x^{(t+1)}}{(\text{mask}*x^{(t+1)})\text{.sum()}}$;\\
cur $\gets$ sample($x^{(t+1)}$);\\
\If{\upshape cur is not None}{
\If{\upshape loop\_back and cur == root\_M}{
    terminate $\gets$ True;\\
}
\Else{
\If{\upshape mask\_attach[cur]}{
M $\gets$ attach(M, molecule(cur));\\
H.child $\gets$ Node(cur);\\
H $\gets$ H.child;
}
\Else {
H $\gets$ H.parent;\\
}
\Else{
\If{\upshape loop\_back}{
return M, root, False;\\
}
\Else{
break;\\
}
}
}
}
}
return M, root, True;\\
Out: molecule, representation of molecule, boolean indicating validity
\end{algorithm2e}

Our implementation in Algorithm \ref{algo2} handles the distinction between revisiting a previous node vs adding a new duplicate of the same motif as a previous node through mask$\_$attach (new nodes which can be attached) vs mask$\_$return (the node which the random walker can backtrack to). This distinction is done by creating duplicates of nodes for each revisit (see \ref{app:B3}). \footnotetext{We concatenate the molecule’s 2048-dimensional morgan fingerprint to the input features. We concatenate the edge-weighted adjacency matrix to the input features.}

We guarantee 100\% validity since we can explicitly check the possible motifs which can be attached to $M$ at each step. When there are neither new motifs to attach, nor existing motifs to return to, the generation terminates with the current M being the final generation output. Please refer to the GitHub for details of the implementation.

\begin{figure}[h!]
  \centering
  \includegraphics[width=0.5\textwidth]{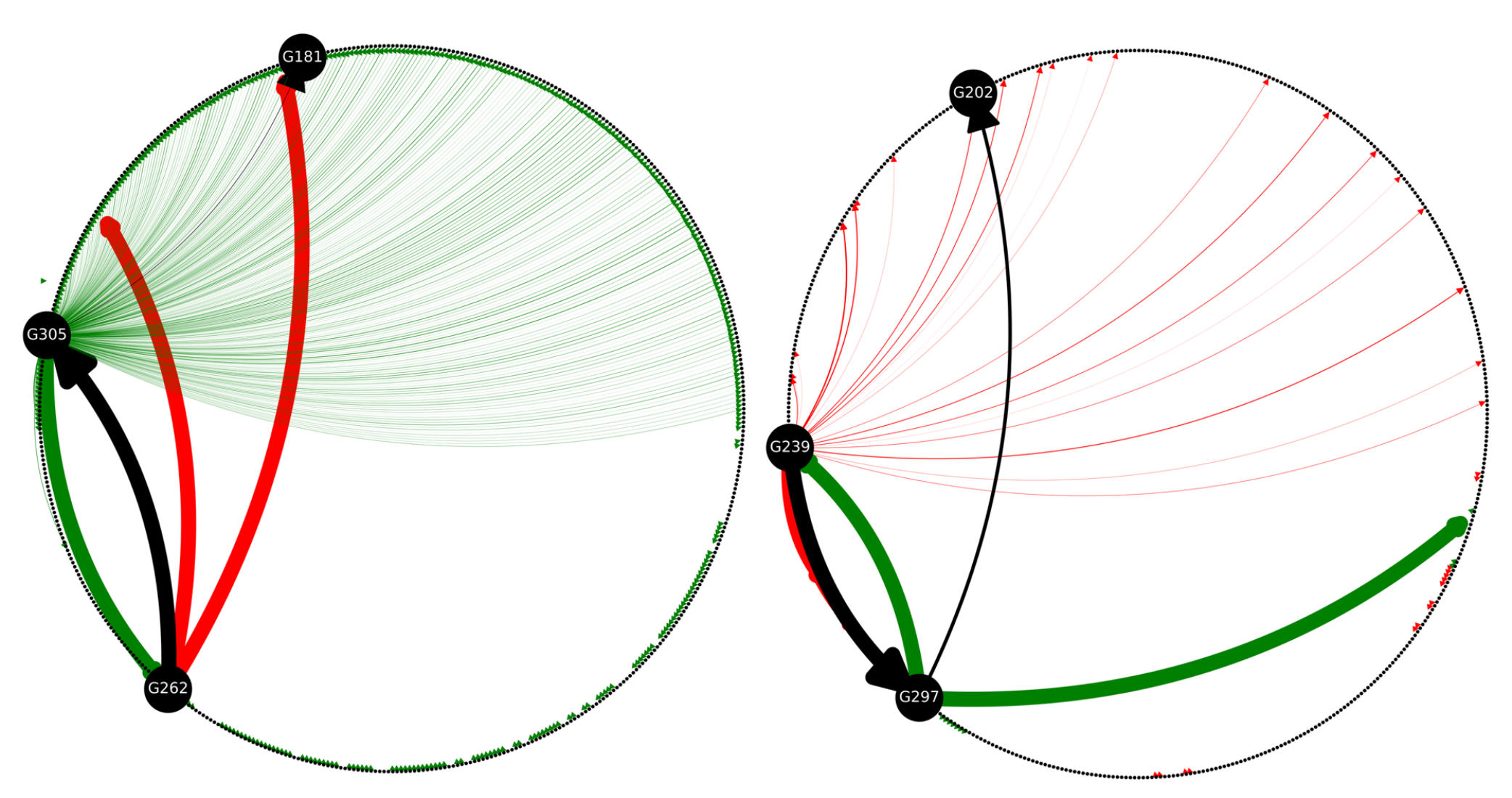}
  \caption{Generation process of novel random walks on HOPV: (Left) $G262 \rightarrow G305 \rightarrow G181$ and (Right) $G239 \rightarrow G297 \rightarrow G202$; The possible transitions from the first and second visited nodes are in \textcolor{red}{Red} and \textcolor{teal}{Green}.}
  \label{fig:soft_hard}
\end{figure}

As shown in Figure \ref{fig:soft_hard}, applying our generation method produces artifacts of learning that invite further scrutiny: “rules" of consecutive motifs. The second example in Figure \ref{fig:soft_hard} shows there are only two possible motifs (green) that can be attached to the $G297$ end of a molecule with the $G239$ and $G297$ functional groups (ignoring the return back to G239, which transitions the state but does not attach a new motif). In the first example, the distribution of possible new motifs to attach to the $G305$ end of a molecule with $G262$ and $G305$ appears more uniform.

\subsection{Extracting Context-Sensitive Grammar Rules} \label{app:F1}
One side product of our generation and verification procedure is the ability to extract “hard” rules. A hard rule is when a certain edge \textit{must} be traversed (probability of 1) under a certain memory and at a certain state. Although our memory is invariant to the order of visited nodes thus far, we search for hard rules by using a best-first algorithm to store all promising trajectories. Table \ref{tab:rules} is a compilation of hard rules learned by our model on the PTC dataset.

\begin{table}[h!]
\centering
\caption{Under our string-based implementation, A[$\rightarrow$B] encodes a random walk trajectory of A$\rightarrow$B$\rightarrow$A. All rules shown are valid, as verified to correspond to valid molecules that can be constructed following the random walk trajectory.}
\label{tab:rules}
\begin{tabular}{|l|l|}
\toprule
Trajectory A                                                                                                                                                                                                                                                                                                                                                                                                                                      & $\Rightarrow$ Trajectory B                                                                                                                                                                                                                                                                                                                                                                                                                                                                                                                                                                                                                                             \\ \midrule
\begin{tabular}[c]{@{}l@{}}{[}'G4'{]}\\ {[}'G27'{]}\\ {[}'G115'{]}\\ {[}'G218'{]}\\ {[}'G283'{]}\\ {[}'G290'{]}\\ {[}'G301'{]}\\ {[}'G335'{]}\\ {[}'G368'{]}\\ {[}'G466'{]}\\ {[}'G272'{]}\\ {[}'G362'{]}\\ {[}'G205'{]}\\ {[}'G435'{]}\\ {[}'G167'{]}\\ {[}'G436'{]}\\ {[}'G224'{]}\\ {[}'G2', 'G4'{]}\\ {[}'G202', 'G205'{]}\\ {[}'G434', 'G435'{]}\\ {[}'G361', 'G362'{]}\\ {[}'G333', 'G393'{]}\\ {[}'G224', 'G225', 'G224:1'{]}\end{tabular} & \begin{tabular}[c]{@{}l@{}}{[}'G4', 'G2'{]}\\ {[}'G27', 'G6'{]}\\ {[}'G115', 'G6'{]}\\ {[}'G218', 'G6'{]}\\ {[}'G283', 'G6'{]}\\ {[}'G290', 'G6'{]}\\ {[}'G301', 'G6'{]}\\ {[}'G335', 'G6'{]}\\ {[}'G368', 'G6'{]}\\ {[}'G466', 'G231'{]}\\ {[}'G272', 'G271'{]}\\ {[}'G362', 'G361'{]}\\ {[}'G205', 'G202'{]}\\ {[}'G435', 'G434'{]}\\ {[}'G167', 'G166'{]}\\ {[}'G436', 'G166'{]}\\ {[}'G224', 'G225'{]}\\ {[}'G2{[}-\textgreater{}G4{]}'{]}\\ {[}'G202{[}-\textgreater{}G205{]}'{]}\\ {[}'G434{[}-\textgreater{}G435{]}'{]}\\ {[}'G361{[}-\textgreater{}G362{]}'{]}\\ {[}'G333', 'G393', 'G333:1'{]}\\ {[}'G224', 'G225{[}-\textgreater{}G224:1{]}'{]}\end{tabular} \\ \bottomrule
\end{tabular}
\end{table}

\section{Detailed Case Study: Harvard Organic Photovoltaic Dataset} \label{app:G}
The Harvard Organic Photovoltaic Dataset (HOPV15) is a comprehensive collection that bridges experimental photovoltaic data with quantum-chemical calculations, serving as a crucial resource in the field of organic photovoltaics. This dataset includes experimental results from literature and corresponding quantum chemical data for a wide range of molecular conformers. These are analyzed using various density functionals and basis sets, including both generalized-gradient approximation and hybrid designs. A key feature of HOPV15 is its utility in calibrating quantum chemical results with experimental observations, aiding in the development of new semi-empirical methods, and benchmarking model chemistries for organic electronic applications. The dataset employs the Scharber model to compute the maximum percent conversion efficiencies for 350 studied molecules, focusing on their HOMO (Highest Occupied Molecular Orbital) values. 

\subsection{Segmentation Strategy} \label{app:G1}
Our segmentation approach involved systematically categorizing molecules based on their functional groups and ring structures. We separated standard functional groups (e.g., vinyl, alcohol) and individual rings (e.g., benzene, thiophene, pyrrole) to understand their unique contributions to photovoltaic properties. Additionally, we paid special attention to complex structures with consecutive rings, acknowledging their impact on the optical and electronic characteristics of the materials. These parameters are impactful to the molecular's HOMO value, which are essential for calculating the open circuit potential and short circuit current density, leading to an understanding of percent conversion efficiency. 

The segmentation strategy is particularly focused on the differentiation and categorization of molecular structures based on their photovoltaic properties and electronic configurations. This includes the separation of standard functional groups such as vinyl, alcohol, ketone, aldehyde, amine, ester, and amide, each identified as individual black fragments. This separation is critical in analyzing their distinct contributions to photovoltaic efficiency and electronic properties.

Moreover, the dataset and segmentation emphasize the unique characteristics of individual rings like benzene, pyrrole, and thiophene by treating them as separate black fragments. This distinction is vital due to their specific Pi-orbital electron delocalization, which plays a crucial role in the photovoltaic properties of the molecules. The segmentation method goes a step further in dealing with complex structures possessing multiple consecutive rings, such as thieno[3,4-b]pyrazine, carbazole, and 2,5-dimethyl-3,6-dioxo-2,3,5,6-tetrahydropyrrolo[3,4-c]pyrrole. These structures are treated as individual entities to accurately reflect their unique HOMO-LUMO bandgaps and electrochemical characteristics, which are central to their functionality in organic photovoltaics.

The segmentation strategy also pays special attention to groups of 2-3 consecutive symmetrical thiophene or pyrrole units, maintaining these as a single black group. This decision is based on the understanding that the electron cloud delocalization across these repeating units significantly influences the optical and electronic properties of the molecules, impacting factors such as light absorption, charge transport, and luminescence. Such approach is essential for advancing the understanding of molecular alignment and stability, thereby optimizing the functional properties of photovoltaic materials.

Meanwhile, all the red group along with the segmented black group are chosen to be either single atom, or closet conjugated rings if the black group is too small (just one or two atoms). This method helps reduce the redundancy and computational resources of the red group. 

\subsection{Heuristic Based Fragmentation} \label{app:G2}
We adopted a heuristic-based, deterministic algorithm to segment molecules across all datasets for our ablation study. Below, we analyze its segmentation quality on the HOPV dataset. We cleave on any bond that satisfies one or either of these conditions: 

\begin{enumerate}[leftmargin=*,topsep=0pt,noitemsep]
    \item Bond connecting two rings.
    \item Bond connecting a ring and an atom with degree greater than 1.
\end{enumerate}

\begin{figure*}[h!]
\centering
\includegraphics[width = 0.8\textwidth ]{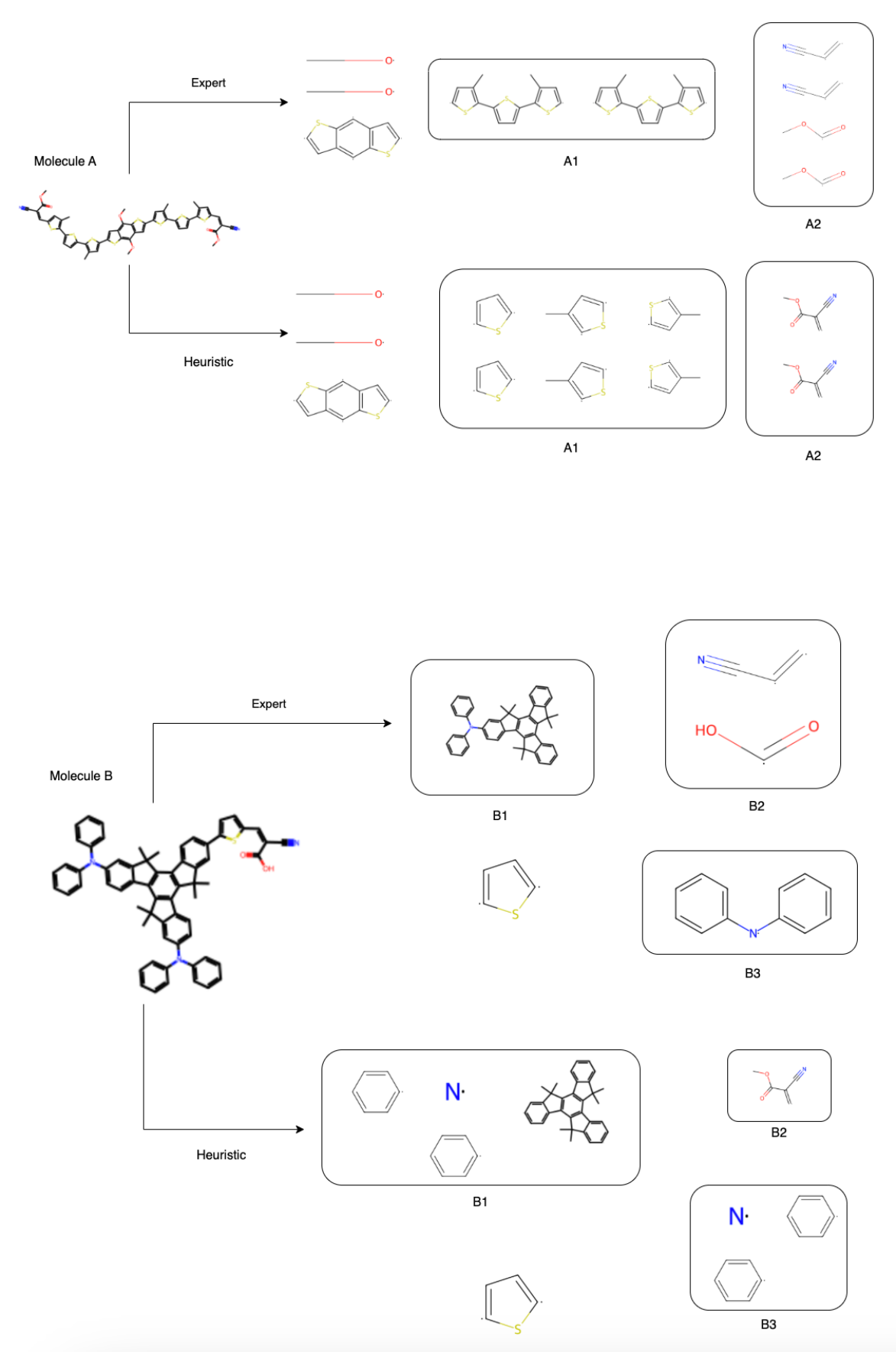}
\caption{We detail the difference between the expert and heuristic segmentations, highlighting how the heuristics are not sufficiently capable. For example, the expert segmentation keeps the 3 thiophene rings together in A1, while the heuristic breaks them up. Similarly, in B1, the expert treats the consecutive rings as one fragment, whereas the heuristic cleaves on bonds connecting them. }
\label{fig:heuristic-expert}
\end{figure*}

This algorithm works for molecules with rings, but tends to not capture functional groups consistently. It either fails to sufficiently segment groups attached to ring like A2 in Figure \ref{fig:heuristic-expert} or cleaves on every ring even when they should be kept together like B1 in Figure \ref{fig:heuristic-expert}.

\subsection{Analysis of Learnt Representations} \label{app:G3}

In this section, we perform an alternate and more accepted means of analysis than the 2D t-SNE analysis done in Section \ref{sec:tsne}. We seek to understand the agreement between our property predictor's learnt representation and the structural similarity over HOPV's test set molecules. Since the final layer embedding is used for prediction, we expect molecules with similar properties to have similar embeddings.

\begin{figure*}[h!]
\centering
\includegraphics[width = \textwidth ]{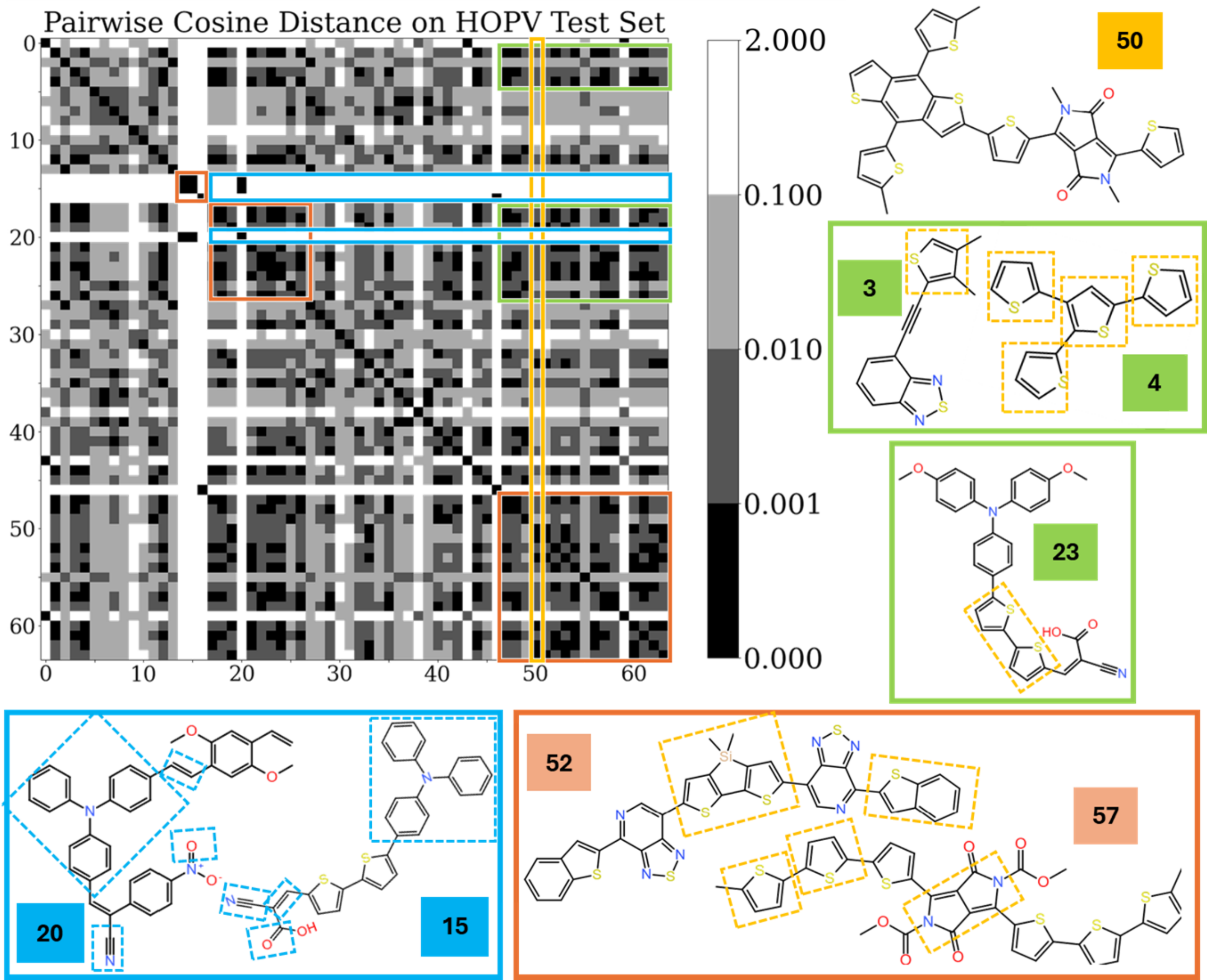}
\caption{There are 64 molecules in this test set indexed from lowest to highest HOMO value. The above grid visualizes the distance between each pair of molecules as a cosine distance between the final layer embeddings of our model, with darker color representing lower distance (higher similarity). We use 4 quantiles, and refer to their ranges as low, medium-low, medium-high, and high similarity.}
\label{fig:hopv-structural}
\end{figure*}

In Figure \ref{fig:hopv-structural}, several groups of trends stand out. Particularly, highlighted in green are cases where the embedding similarity is high despite dissimilar HOMO property values; blue marks cases where the embedding similarity is low, and red marks sections that are similar in property, structure, and embedding. We detail each of these, basing comparison against molecule 50 for illustration:

\begin{itemize}[leftmargin=*,topsep=0pt,noitemsep]
    \item For the topmost green section, molecules in the range 1-4 have similar components as those with higher HOMO values, though are much smaller in size and relatively disordered. For instance, molecules 3 and 4 each share key subcomponents (thiophene groups) with molecule 50, despite having quite different overall structure. The embedding similarity between (50, 4) and (50, 3) is thus medium-low and medium-high.
    \item For the red sections along the diagonal, molecules in the ranges 14-16 and 17-26 cluster together. These tend to have an over-representation of electron-withdrawing groups in in non-symmetric locations in the structure, particularly methoxy, cyano, and carbonyl groups, without sufficient electron donating groups. Molecules 15 and 20 are shown as examples, and their embedding similarity is high. Blue outlines mark similar sub-groups between 15 and 20.
    \item For the second-from-top green section, we again consider molecules in the range 18-26, where they show high similarity to the highest band in the range 47-63. These share many component structures, for instance thiophenes groups and derivatives. Molecule 23 is shown as an example, and has a barbituric acid core on one side, an electron withdrawing group, with methoxy groups on benzene rings on the other side, with a nitrogen atom between benzene rings, contributing to electron delocalization. The most likely explanation is that similar high-sterics groups have developed similar embeddings in this case.
    \item For the bottom-right red section, molecules in the range 47-63 generally cluster together, reflecting the model's ability to agree on both structural and property similarity. They tend to have an alternating pattern of electron-donating and electron-withdrawing groups which can increase the HOMO and provide a more direct pathway for charge transport. Yellow outlines mark matching and similar groups with molecule 50. In these cases, more than simply thiophene shows similar or the same structure. The embedding similarity between (50, 52), (50, 57), (52, 57) are all medium-high.
\end{itemize}
These insights show how complex molecule structure affects the measured property in this application, and how both structure and property are captured in the embedding. We hope the analysis provides more insights into how structural priors in our representation facilitates learning and generalization.

